\documentclass[sigconf]{acmart}
\renewcommand\footnotetextcopyrightpermission[1]{}
\AtBeginDocument{%
  }

\setcopyright{acmlicensed}
\copyrightyear{2018}
\acmYear{2018}
\acmDOI{XXXXXXX.XXXXXXX}
\acmConference[Conference acronym 'XX]{Make sure to enter the correct
  conference title from your rights confirmation email}{June 03--05,
  2018}{Woodstock, NY}
\acmISBN{978-1-4503-XXXX-X/2018/06}

\usepackage{amsthm}

\usepackage{amssymb}
\usepackage{amsmath}  
\usepackage{braket}   
\usepackage{booktabs}
\usepackage{bm}
\usepackage{colortbl}
\usepackage{multirow}
\usepackage{graphicx}
\usepackage{float}
\usepackage{subcaption}
\usepackage{xspace}
\usepackage{algorithm}
\usepackage{algorithmic}

\newtheorem{lemma_app}{Lemma}

\newtheorem{definition}{Definition}
\newtheorem{proposition}{Proposition}
\newcommand{\ie}{\emph{i.e.,}\xspace}

\newcommand{\eg}{\emph{e.g.,}\xspace}

\begin{document}

\title{Continuous-Time Quantum Walks based Graph Neural Network}

\author{Yuliang Zhan}
\email{zhanyuliang@ruc.com}
\affiliation{%
  \institution{Renmin University of China}
  \city{Beijing}
  \country{China}
}
\author{Zefeng Gao}
\email{zfgao@ruc.com}
\affiliation{%
  \institution{Renmin University of China}
  \city{Beijing}
  \country{China}
}

\author{Jian Li}
\email{lijian2022@ruc.edu.cn}
\authornote{Corresponding author.}
\affiliation{%
  \institution{Renmin University of China}
  \city{Beijing}
  \country{China}
}

\author{Yang Liu}
\email{Liuyang22@ucas.ac.cn}
\affiliation{%
  \institution{Renmin University of China}
  \city{Beijing}
  \country{China}
}

\author{Hao sun}
\email{haosun@ruc.edu.cn}
\authornotemark[1]
\affiliation{%
  \institution{Renmin University of China}
  \city{Beijing}
  \country{China}}

\renewcommand{\shortauthors}{Trovato et al.}

\begin{abstract}
Graph Neural Networks (GNNs) are widely used on graph-structured data. However, most GNNs suffer from two key weaknesses. First, message passing acts as a low-pass filter based on the homophily assumption, so it performs poorly on heterophilic graphs, where connected nodes differ. Second, stacking layers makes node features converge exponentially to constants, a problem known as over-smoothing. Existing work usually addresses the two weaknesses separately. The few methods that target both rely on empirical heuristics, and many over-smoothing solutions further sacrifice the expressive power of the model.
To address both weaknesses with theoretical justification, we propose \textbf{CTQW-GNN}, a GNN built on Continuous-Time Quantum Walks (CTQW). The design is motivated by two properties of the CTQW propagator $e^{-\mathrm{i}Ht}$. (i) It is unitary and its eigenvalues lie on the unit circle, so no frequency component is damped. This directly counters the low-pass bias. (ii) Unitarity preserves the feature norm, so the Dirichlet energy does not decay exponentially with depth. This directly counters over-smoothing.
Guided by these properties, CTQW-GNN combines three aggregation modules, each motivated by a specific gap in prior work. \textit{CTQW-based Aggregation} evolves node features through the unitary propagator. It captures the mid- and high-frequency signals needed for heterophilic graphs and provably keeps the Dirichlet energy from collapsing, so it addresses both weaknesses in a single branch. \textit{CTQW-Attention Aggregation} builds a multi-hop neighbor graph from the CTQW amplitudes and applies attention over it, so distant homophilic nodes that single-hop aggregation misses can still be reached. \textit{LF Aggregation} is a standard low-pass branch (GAT) that preserves accuracy on strongly homophilic graphs, where a pure CTQW branch is suboptimal. We further provide a spectral-gap analysis that explains the energy preservation, and a Lieb--Robinson-type bound that gives a principled rule for choosing the walk-time $t$.
Experiments on $14$ benchmarks ($9$ heterophilic, $5$ homophilic) show that CTQW-GNN reaches state-of-the-art accuracy on every dataset. Krylov propagation and Chebyshev-based sparsification keep the CTQW-related computation linear in the number of edges under the stated sparsification rule.

\end{abstract}

\ccsdesc[500]{Computing methodologies~Machine learning}
\keywords{Deep Graph Neural Networks, Over-smoothing, Heterophilic Graph, Continuous-Time Quantum Walks}


\maketitle

\section{Introduction}

Graph Neural Networks (GNNs) have seen widespread application across numerous important fields, such as quantum physics~\cite{yu2023efficient}, traffic networks~\cite{liu2021self} and recommendation systems~\cite{quan2023robust}.
This is primarily attributed to the capability of GNN to capture node features and graph topology information.
Despite these significant advancements~\cite{gao2023addressing,tu2024adaptive}, most of existing GNN models still have two inherent weaknesses (\ie~homophily assumption and over-smoothing) that can diminish their performance. 

The first weakness of GNNs is the homophily assumption, as they update node or edge features by aggregating information from neighbors. This aggregation method acts as a low-pass filter~\cite{wu2019simplifying,li2019label}, which preserves similarities and filters out differences among connected nodes, thereby promoting feature uniformity~\cite{bo2021beyond}. However, recent studies indicate that while this filtering mechanism excels in homophilic graphs with similar connected nodes, it reduces GNNs performance in learning node representations on heterophilic graph where connected nodes differ (heterophily problem)~\cite{zhu2020beyond,li2024pc}. To improve GNN performance on heterophilic graphs, many researchers are attempting to construct high-frequency filters to capture high-frequency information~\cite{bo2021beyond,maskey2024fractional}. This approach is motivated by findings that high-frequency information is useful for learning on heterophilic graph~\cite{bo2021beyond}. Another prevalent approach attempt involves aggregating multi-hop neighbor nodes to capture long-range homophilic dependencies~\cite{zheng2023finding,liang2024predicting}. 

The second weakness of GNNs is the over-smoothing problem where increasing the number of layers in GNN causes the features of all nodes exponentially converge towards constant values~\cite{10.5555/3504035.3504468, 10.1145/3583780.3614997}. This leads to most applied GNNs being shallow, limiting their capabilities. Recently, researchers have primarily focused on mitigating over-smoothing by implementing drop operation~\cite{rong2019dropedge}, normalization~\cite{zhou2020towards,zhou2021understanding}, and modifying the dynamical systems of GNNs~\cite{rusch2022graph,wang2023acmp}. However, previous study posits that while some methods can mitigate over-smoothing, this comes at the cost of sacrificing the expressive performance of GNNs~\cite{rusch2023survey}.

Heterophily and over-smoothing are often studied as separate problems~\cite{pmlr-v139-chamberlain21a,eliasof2021pde}, but recent evidence suggests that they are closely related~\cite{yan2022two}. Recent works have found that methods addressing heterophily can also alleviate over-smoothing~\cite{chien2021adaptive, eliasof2024feature}, and vice versa~\cite{rusch2022graph,wang2023acmp}. These methods, however, rely on empirical observation rather than theoretical guarantees. A reliable solution requires an aggregation method that captures diverse frequencies and expands node neighborhoods while \emph{provably} preventing the node features from converging to constants across many layers.


In this paper, we propose \underline{C}ontinuous-\underline{T}ime \underline{Q}uantum \underline{W}alks based on \underline{G}raph \underline{N}eural \underline{N}etwork, namely \textbf{CTQW-GNN}, to theoretically addressing both weaknesses. We first derive multi-hop node connectivity and edge weights through CTQW. Then, we utilize one-hop connectivity to aggregate low-frequency information and multi-hop connectivity to capture information from distant nodes. This enables CTQW-GNN to effectively learning in both homophilic and heterophilic graphs. Specifically, we adopt the attention mechanism from Graph Transformer~\cite{shi2020masked} to aggregate distant nodes via CTQW-induced multi-hop connectivity, thereby capturing long-range homophilic relationships. Note that previous study points that most graphs are not purely homophilic or heterophilic, but rather fall somewhere in between~\cite{mao2024demystifying}. Furthermore, mid-frequency information has been shown to enhance GNN performance in such mixed-pattern graphs~\cite{maskey2024fractional}. CTQW-derived edge weights naturally enable the aggregation of mid- and high-frequency information, so we can use CTQW effectively capturing mid-range spectral components in the graph~\cite{6130444}. 
Finally, we combine the information aggregated from one-hop and multi-hop connectivity with that aggregated through CTQW-derived edge weights. This combined feature serves as the input for the next layer of CTQW-GNN. Since the norm-preserving property of CTQW, the Dirichlet energy of graph does not exponentially converge to zero, preventing over-smoothing in the CTQW-GNN model. 
The contributions of this paper are three fold:

$\bullet$ \textbf{CTQW-inspired aggregations.} We design CTQW-based and CTQW-Attention aggregations that preserve quantum superposition and phase-driven interference.

$\bullet$ \textbf{CTQW-GNN with provable guarantees.} The model combines three aggregations to capture low-, mid-/high-frequency, and long-range information. We provide a spectral-gap analysis and a Lieb--Robinson bound $r_{\mathrm{eff}}(t)\!=\!2\lambda_{\max}t/\pi$ that together explain why CTQW avoids exponential energy decay and how to choose $t$.

$\bullet$ \textbf{Extensive experiments.} CTQW-GNN attains state-of-the-art accuracy on all 14 benchmarks, outperforming strong baselines on every dataset ($+1.07\%$ on average; $+1.06$--$3.31\%$ on saturated homophilic datasets).

\section{Related Work}
{\textbf{Heterophilic GNN.}}
To address the heterophily problem, many works focused on designing Heterophilic GNNs without the homophily assumption~\cite{li2022finding,liang2024predicting}. These works can mainly be divided into two types: the first type of methods combined high-pass and low-pass filters to capture information from neighboring nodes~\cite{pmlr-v202-guo23i,tu2024adaptive}, since the high-frequency information was helpful to address the heterophily problem~\cite{bo2021beyond}. 
The second type of methods expanded the neighborhood of nodes to aggregate distant homophilic information~\cite{li2022finding,liang2024predicting}. Our model distinguishes from existing methods by offering a hybrid integration of approaches. It adeptly utilizes classic aggregation methods, such as Graph Convolutional Networks (GCN), for the aggregation of low-frequency information. Meanwhile, for mid-range and high-frequency information aggregation, it employs the advanced CTQW strategy, thereby combining the best of both parts to enhance performance and efficiency. Furthermore, our model leverages the connectivity derived from CTQW to adeptly capture long-range homophilic dependencies. This innovative application of CTQW significantly enhances the expressive power of GNNs, allowing for a more nuanced and comprehensive representation of complex graph structures.

{\textbf{Over-smoothing.}}
Over-smoothing is a well-known problem in GNNs, characterized by node features exponentially converging towards a same constant value as the number of layers increases. Numerous methods have been proposed to mitigate it mainly from Drop Operation~\cite{rong2019dropedge}, Normalization~\cite{zhou2021understanding}, and modifying the dynamical systems of GNNs~\cite{rusch2022graph}. Despite notable progress, there is a trade-off in existing methods that can limit GNN expressiveness~\cite{rusch2023survey}. We introduce CTQW-based aggregation to counteract over-smoothing. Unlike most research that treats over-smoothing and heterophily separately, our model tackles both issues concurrently, enhancing GNN performance on complex graph tasks. Furthermore, we substantiate the reliability of this approach theoretically.

\section{Preliminary}
\paragraph{Notations}
We define an undirected graph $\mathcal{G}=(\mathcal{V},\mathcal{E})$, with $\mathcal{V}$ as the node set of size $N$ and $\mathcal{E}$ as the edge set. The adjacency matrix is $A \in \mathbb{R}^{N \times N}$. The degree matrix $D$ is diagonal with $D_{ii}=\sum_j{A_{ij}}$. The normalized graph Laplacian $L=I-D^{-\frac{1}{2}}AD^{-\frac{1}{2}}$ (where $I$ is the identity matrix) is symmetric and expressed as $U\Lambda U^T$. Here, $\Lambda = diag([\lambda_1,\lambda_2,...,\lambda_N])$ represents graph signal frequencies and $U={\{ u_i\} }_{i=1}^{N}$ denotes the frequency components.

\paragraph{Graph Fourier Transform}
We treat $U$ as the base in graph Fourier transform. The transform of a graph signal $x \in \mathbb{R}^n$ on $\mathcal{G}$ is $\hat{x} = U^T x$, and the inverse is $x = U\hat{x}$. The convolution of the graph signal $x$ with the kernel $f$ is:
\begin{equation}
    \label{eq:gnn}
    (f*x)_{\mathcal{G}} = U((U^Tf) \odot (U^Tx)) = Ug_{\theta}U^Tx.
\end{equation}
where $\odot$ denotes the hadamard product, and $g_{\theta}$ is a learnable filter which can adjust the frequency response of the graph signal.For example, GCN defines the convolutional kernel $g_{\theta} = I - \Lambda$, where $\lambda_{g_{\theta},i} = 1-\lambda_i$. It shows that the convolutional kernel of GCN is low-pass filter.
\paragraph{Over-smoothing and Dirichlet Energy}
Recent literature mainly employ graph Dirichlet energy (DE) to measure the similarity of features between nodes, thereby defining over-smoothing\cite{rusch2023survey}. The \textit{Dirichlet energy} $E$ defined on an undirected graph $\mathcal{G}$ with node features $X$ is :
\begin{equation}
    \label{eq:DE}
    E(X)=\frac{1}{N}\sum_{i \in \mathcal{V}}{\sum_{j \in \mathcal{N}_i} \left\Vert X_i-X_j \right\Vert ^2_2}
\end{equation}
With Dirichlet energy, we can define over-smoothing like~\cite{rusch2022graph} as following:
\begin{definition}
    \label{def:oversmoothing}
    Let $X^n$ denote the node feature at the $n$-th layer of GNN. Over-smoothing is defined as the exponential convergence to zero of the layer-wise Dirichlet energy as a function of n:
    \begin{equation}
        E(X^n) \leq ae^{-bn},
    \end{equation}
    where $a$ and $b$ are constants and $a,b>0$
\end{definition}
In other words, the node features will exponentially converge to a constant value as the number of layers increases.

\paragraph{Quantum Walks: a Primer}
A \emph{classical random walk} (CRW) on a graph propagates a probability vector $p(t)\!\in\!\mathbb{R}^N_{\ge 0}$ by the diffusion equation $\dot{p}=-Lp$ with solution $p(t)=e^{-Lt}p(0)$~\cite{chung1997spectral}. A \emph{quantum walk} (QW), introduced by~\citet{aharonov1993quantum} and surveyed by~\citet{kempe2002quantum}, replaces real probabilities with complex amplitudes $\psi(t)\!\in\!\mathbb{C}^N$ and the dissipative generator $-L$ with a Hermitian Hamiltonian $H$. The walk evolves by the Schr\"odinger equation $\mathrm{i}\dot{\psi}=H\psi$, whose solution $\psi(t)=e^{-\mathrm{i}Ht}\psi(0)$ is \emph{unitary}: it preserves the $\ell_2$ norm of $\psi$ rather than the $\ell_1$ norm of $p$. Its continuous-time variant was developed by~\citet{farhi1998quantum}. Two intuitive advantages over CRW motivate our use: (i) eigenmodes acquire \emph{different phases} $e^{-\mathrm{i}\lambda_l t}$ instead of different decay rates $e^{-\lambda_l t}$, so high-frequency content is preserved (Section~\ref{para:spectralgap}); and (ii) phase-coherent superposition spreads the walker \emph{ballistically} (distance $\propto t$) rather than \emph{diffusively} (distance $\propto\!\sqrt{t}$), yielding exponential separation in hitting time on certain graphs~\cite{farhi1998quantum,childs2002example}. We use QW purely as a unitary linear operator on $\mathbb{C}^N$, classically simulated (Section~\ref{method}); no quantum hardware is required.
\begin{figure*}[ht!]

    \centering
    \small
    \includegraphics[width=\textwidth]{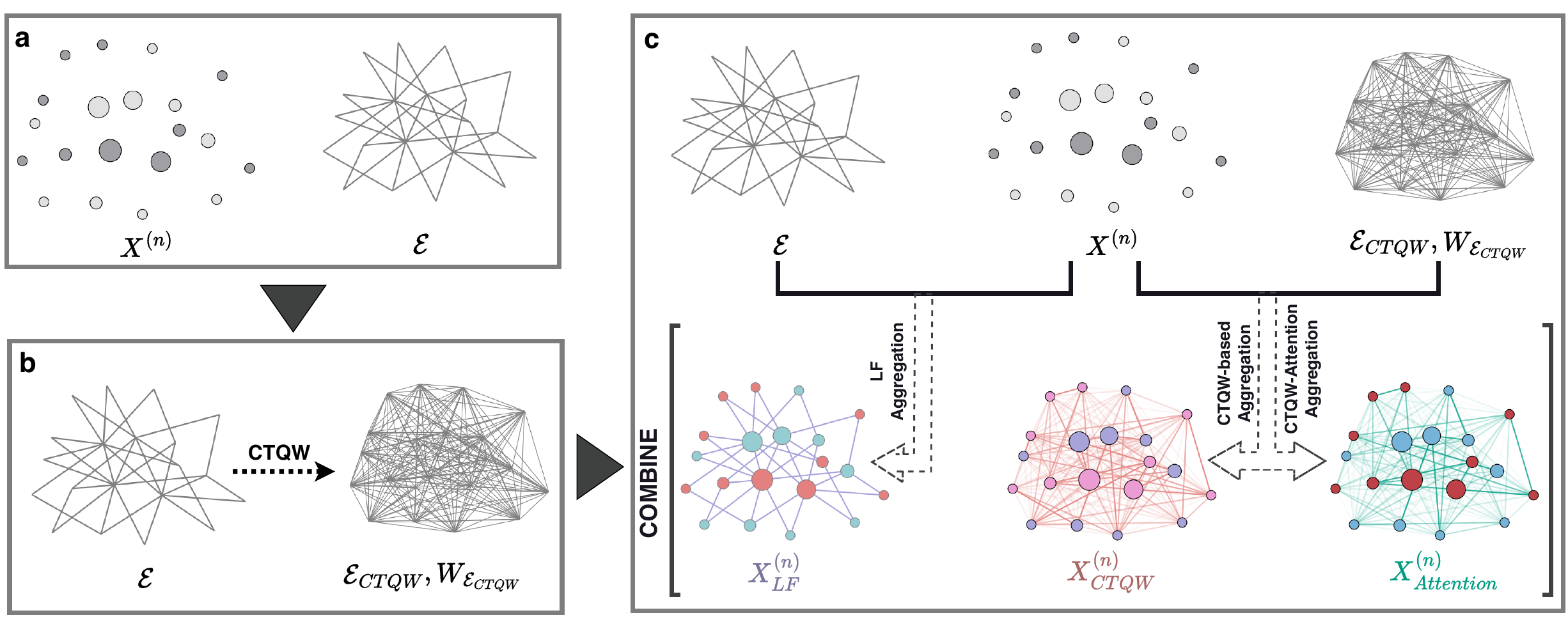}
    \caption{The overview of CTQW-GNN for heterophilic graphs and over-smoothing problem. \textbf{(a).} We define an undirected graph $\mathcal{G}=(\mathcal{V},\mathcal{E})$, with $\mathcal{V}$ as the node set of size $N$ and $\mathcal{E}$ as the edge set. \textbf{(b).} We conduct a CTQW on the original graph $\mathcal{G}$ to obtain a new edge set $\mathcal{E}_{CTQW}$ with edge weights $W_{\mathcal{E}_{CTQW}}$ ; both connectivity relations $\mathcal{E}$ and $\mathcal{E}_{CTQW}$ share the same node features $X^{(n)}$. \textbf{(c).} We employ three distinct information aggregation methods proposed in this paper to derive three types of node information: $X^{(n)}_{LF}$, $X^{(n)}_{CTQW}$ and $X^{(n)}_{Attention}$. through the three types of features, we obtain the node features $X^{(n+1)}$ for the next layer. The details of the formula are provided in Section~\ref{method}.}
    \label{fig-main}
\end{figure*}

\paragraph{Homophily Ratio}
We use a single term \emph{homophily}—the tendency of connected nodes to share the same label—throughout the paper; the words ``homogeneity'' and ``heterogeneity'' are avoided to prevent confusion with the graph-theoretic notion of node/edge type homogeneity (in which all nodes/edges share a single type)~\cite{shi2016survey}. We measure homophily at the graph and node level.
\begin{definition}[Graph Homophily Ratio]
    \label{def:homophily_graph}
    Let $C \in \mathbb{R}^{\left| y\right| \times \left| y\right|}$ where $\left| y\right|$ is the number of node categories and $C_{ij}=|\{(u,v):(u,v) \in \mathcal{E} \wedge y_u = i \wedge y_v = j\}|$. The \textit{Graph Homophily Ratio} is
    \begin{equation}
        h = \frac{\sum_{i}^{N}\sum_j^{i==j}{C_{ij}}}{\sum_{i}^{N}\sum_j^{i \neq j}{C_{ij}}}.
    \end{equation}
\end{definition}
Homophilic graphs have a high $h$; heterophilic graphs have a low $h$. The node-level analogue measures the local pattern around each node:
\begin{definition}[Node Homophily Ratio]
    \label{def:homophily_node}
    Let $y \in \mathbb{R}^N$ be the node labels and $\mathcal{N}(v_i)$ the neighborhood of $v_i$ with degree $d_{v_i}$. The \textit{Node Homophily Ratio} is
    \begin{equation}
        h_{v_i} = \frac{\left| \{u \in \mathcal{N}(v_i):y_u = y_{v_i}\} \right|}{d_{v_i}}.
    \end{equation}
\end{definition}

\section{Method}
\label{method}
In this section, we describe CTQW-GNN. We first outline our approach, then introduce the details of CTQW-based Aggregation and CTQW-Attention Aggregation. Finally, we present how the model in this paper integrates the three methods of information aggregation and justify why our approach can mitigate over-smoothing and enhance the ability to handle heterophilic graphs.

\subsection{Overview}
Existing methods typically address Over-smoothing and heterophily problem as separate problem. To address both problems concurrently, we further augment the capability of model to aggregate information by employing a novel method of information aggregation inspired by CTQW. 
This approach helps prevent the aggregated node information from exponentially converging to a constant value as the number of layers increases, thereby avoiding over-smoothing. Forthemore, the novel aggregation method can aggregate information from various frequency and, based on the higher-order neighbor connectivity information derived from CTQW, aggregate information from distant neighbors, thus improving performance on heterophilic graphs.


To achieve this goal, we first get a new graph along with the weights of edges by CTQW. Based on the node connection weights of the new graph, \textit{\textbf{CTQW-based Aggregation}} can be performed. This method aggregates mid-range and high-frequency information to improve performance on heterophilic graphs. Furthermore, due to the unitary norm preservation of CTQW, the CTQW-based branch provides a non-decaying energy component that mitigates over-smoothing during information aggregation.
Subsequently, we use the new graph obtained by CTQW to aggregate higher-order neighbor information broadens the neighborhood size, thereby capturing more homophilic node. To find homophilic nodes among a large set of neighbors, we employs the attention mechanism of Graph Transformers during the aggregation of higher-order neighbors; hence, this method is referred to as \textit{\textbf{CTQW-Attention Aggregation}}. 
Furthermore, to ensure that the model maintains strong performance on homophilic graphs, this paper also utilizes a low-pass GNN to aggregate homophilic information. This method is termed \textbf{L}ow-\textbf{F}requency \textbf{Aggregation} (\textit{\textbf{LF Aggregation}}). An overview of our approach is depicted in Figure~\ref{fig-main}.
The three aggregators address distinct failure modes of message passing.
(1) \emph{Over-smoothing}: damping $e^{-\lambda_l t}$ is replaced by the unitary multiplier $e^{-\mathrm{i}\lambda_l t}$ on the unit circle, so the Dirichlet energy of the $X_{CTQW}$ branch is layer-invariant (Prop.~\ref{proposition:oversmoothing_CTQW}, Lemma~\ref{lemma:spectralgapbound}).
(2) \emph{Heterophily}: the same eigenmode preservation gives CTQW-based Aggregation an all-pass filter, while the $\epsilon$-thresholded $|e^{-\mathrm{i}Ht}|$ exposes a Lieb--Robinson-bounded long-range graph $\mathcal{G}_{CTQW}$ of radius $r_{\mathrm{eff}}(t)$.
(3) \emph{Saturated homophilic graphs}: an LF branch (GAT) handles regimes where low-pass aggregation is already optimal. Concatenating the three branches (Eq.~\ref{eq:feature}) ensures one branch's failure cannot collapse the representation (Prop.~\ref{proposition:oversmoothing_detail}).

    

\subsection{CTQW-based Aggregation}

{\textbf{Continuous Time Quantum Walks.}}
The CTQW on graph $\mathcal{G}$ represents the evolution of node states in an $N$-dimensional Hilbert space with orthonormal basis ${\ket {a}}$ where $a=1,2,3,...,N$ and $ \braket{a|b}=\delta_{ab}$, with $\delta_{ab}=1$ if and only if $a=b$, and $0$ otherwise. Node states $\ket{\psi(t)}$ at time $t$ evolve according to the Schrödinger equation:
\begin{equation}
    \label{eq:CTQW_ode}
    \mathrm{i} \hbar \frac{d}{dt} \ket {\psi (t)} = H \ket {\psi (t)},
\end{equation}
where $H$ is the Hamiltonian matrix of the CTQW system and $\mathrm{i}$ is the imaginary unit, which is different from the previous equation $i$. $\hbar$ is Planck's constant, which is commonly considered to be 1. The state of nodes evolves from the initial state $\ket{\psi(0)}$ as:
\begin{equation}
    \label{eq:CTQW}
    \ket {\psi (t)} = e^{-\mathrm{i}Ht} \ket {\psi (0)},
\end{equation}
Previous works show that a finite graph is given where classical and quantum walks give an exponential separation in expected hitting time~\cite{farhi1998quantum, childs2002example}. Continuous-Time Random Walk (CTRW, \ie $x(t) = e^{-Lt} x(0)$, where $L$ denotes the graph Laplacian matrix.) is a diffusion process and serves as the primary information propagation mechanism in most existing homophilic GNN. 

\begin{figure}[t!]    
    \centering
    \begin{subfigure}[b]{0.23\textwidth}
        \centering
        \includegraphics[width=\linewidth]{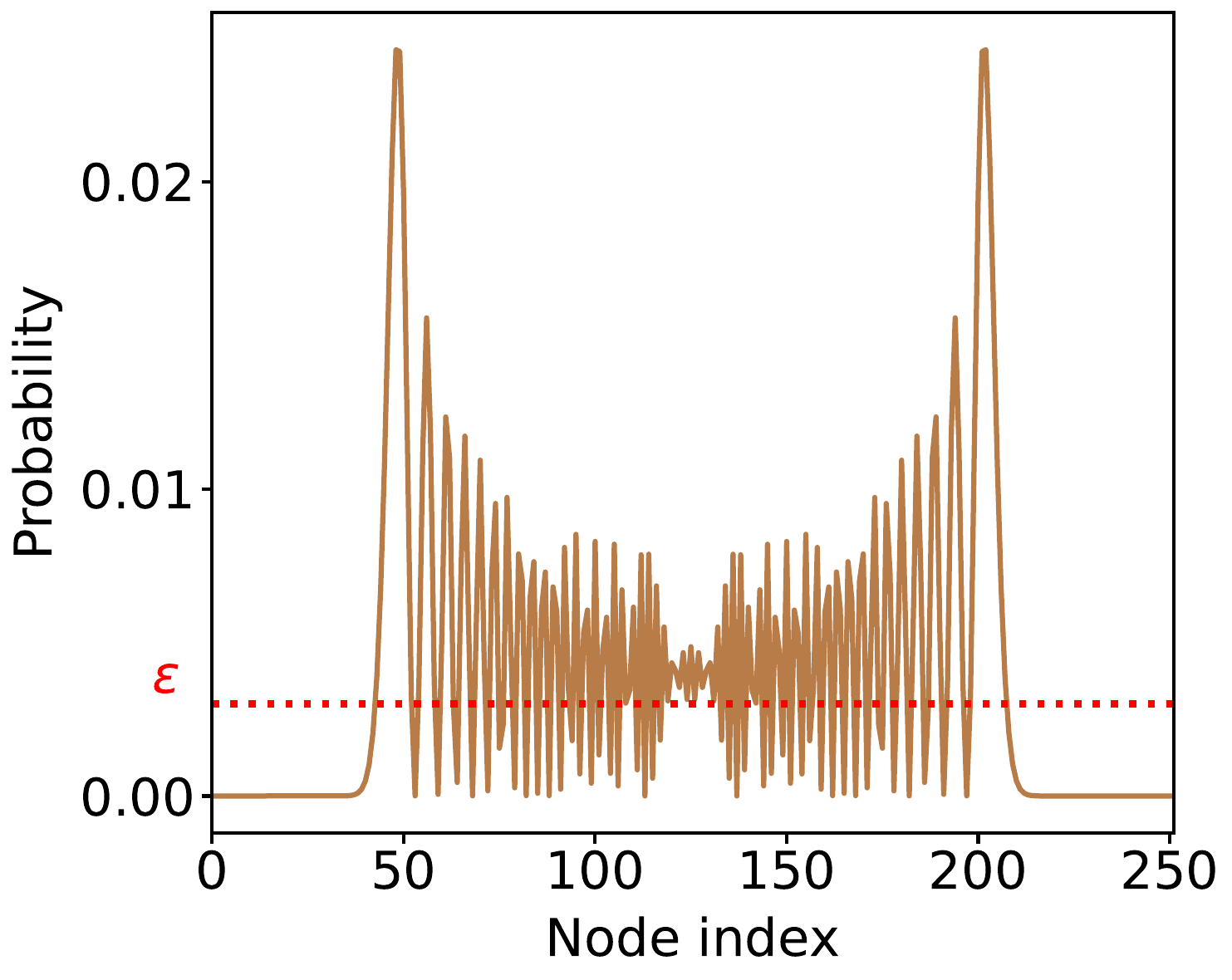}
        \caption{CTQW: Interference-Driven Spread with Sharp Peaks.}
        \label{fig:distribution_CTQW}
    \end{subfigure}
    \hfill
    \begin{subfigure}[b]{0.23\textwidth}
        \centering
        \includegraphics[width=\textwidth]{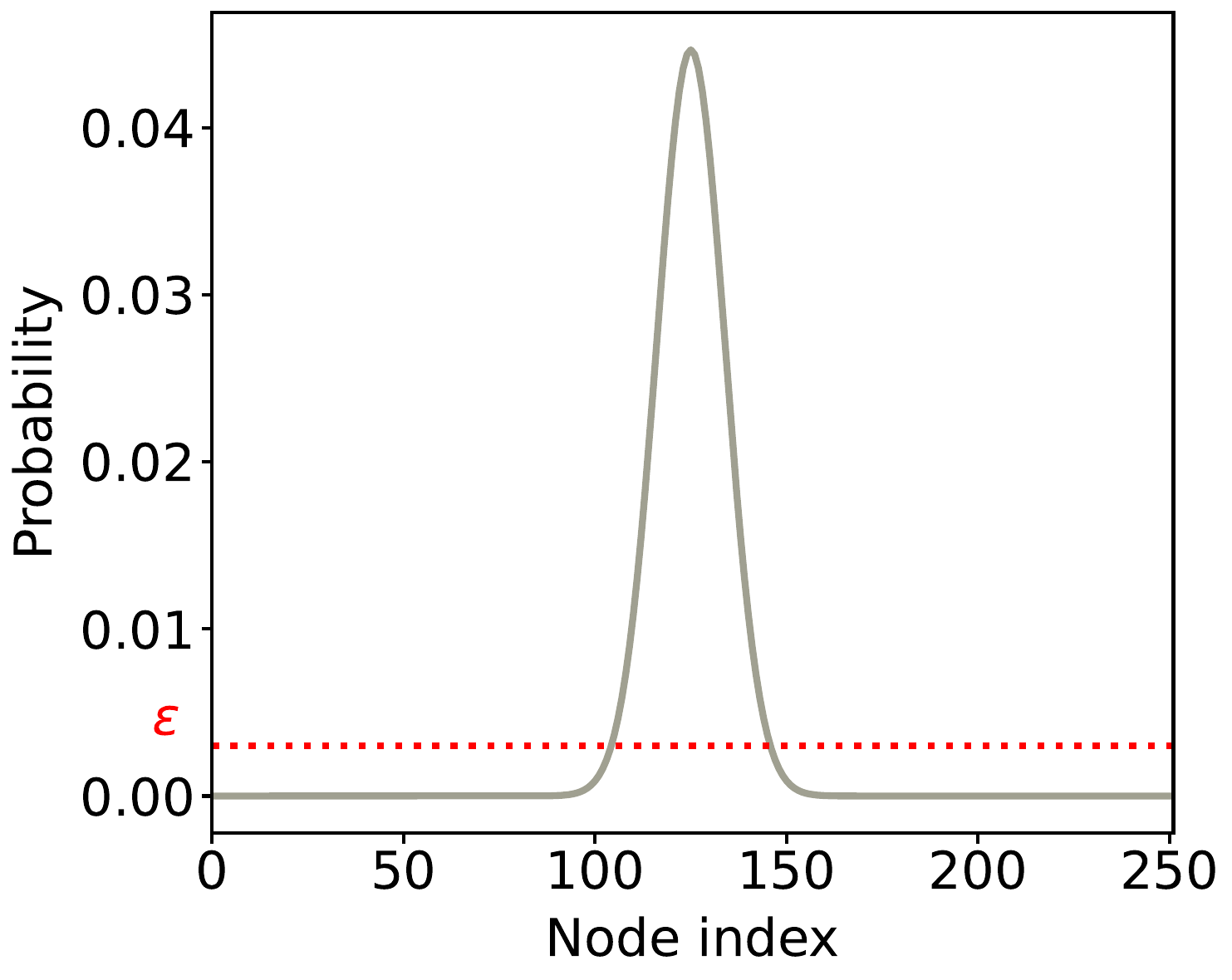}
        \caption{CTRW: Classical Diffusion Centered Near Origin.}
        \label{fig:distribution_CTRW}
    \end{subfigure}
    \caption{Probability Distribution Comparison Between CTQW and CTRW on a Path (starting from node 126, $t=40$).}
    \label{fig:distribution}
\end{figure}

Figure~\ref{fig:distribution} illustrates the diffusion behavior and probability distribution of quantum walks and classical random walks on the same graph structure (a path graph) under identical time conditions. We can see that under the same time duration, the propagation range of CTQW is significantly larger than that of classical CTRW. 
Furthemore, it can be intuitively observed that CTRW-based information aggregation tends to assign higher weights to nearby nodes, which will lead to the over-smoothing. In contrast, CTQW places more emphasis on both higher-order neighboring nodes and mid-order neighboring nodes. This observation motivates us to leverage CTQW as the foundation for designing an information propagation mechanism that mitigates over-smoothing while effectively addressing heterophily.

{\textbf{Aggregation via CTQW. }}
We propose CTQW-based Aggregation, a novel aggregation method based on Eq.~\ref{eq:CTQW}. Different from the original CTQW, in order to accommodate the multi-layer structure of GNNs, the CTQW-based Aggregation discretizes time, evolving with layers rather than with continuous time. The evolution of node features is governed by the following equation:
\begin{equation}
    \label{eq:CTQW_aggregation}
    X_{CTQW}^{(n)} = f(e^{-\mathrm{i}Ht} X^{(n)}),
\end{equation}
where $X^{(n)}$ and $X_{CTQW}^{(n)}$ are the feature of nodes at the $n$-th layer of the GNN and the output of $n$-th layer of CTQW-based Aggregation, respectively. The $t$ in $e^{-\mathrm{i}Ht}$ is a hyperparameter that influences the range of neighbors obtained through the quantum walk. The function $f(\cdot)$ concatenates the imaginary and real parts of the input.

\textbf{Choice of Hamiltonian $H$. }
Either the adjacency matrix $A$ or the normalised Laplacian $L=I-D^{-1/2}AD^{-1/2}$ is a valid Hermitian Hamiltonian, related by an eigenvalue translation ($\lambda^L_l=1-\lambda^A_l/d$ on $d$-regular graphs), so they generate the same family of unitaries up to a global phase. We use $A$ by default for three reasons: (i) $A$ has the lowest constant in $\mathrm{nnz}$, minimising every Krylov mat-vec; (ii) the trivial mode of $L$ is the constant ``smoothing'' direction $\lambda\!=\!0$, which we wish to avoid privileging; (iii) $A$ matches the tight-binding Hamiltonian in solid-state physics~\cite{bose2003quantum}. 

\textbf{Practical Simulation of $e^{-\mathrm{i}Ht}X^{(n)}$. }
A natural concern is whether the quantum state $\ket{\psi(0)}$ in Eq.~\ref{eq:CTQW} entails a $2^N$-dimensional Hilbert space and consequently requires exponential-time simulation. This holds for general many-body quantum systems but \emph{not} for single-particle CTQW: the walker lives in the $N$-dimensional vertex space $\mathbb{C}^N$ (one amplitude per node), not in the $2^N$ Fock space of $N$ qubits. Hence $e^{-\mathrm{i}Ht}$ is an $N\!\times\!N$ unitary that we simulate classically with no quantum hardware. We never materialise the dense matrix; for each feature column $X^{(n)}_{:,k}\!\in\!\mathbb{R}^N$ we apply $e^{-\mathrm{i}Ht}$ via a $k_\text{iter}\!\le\!20$-step Lanczos Krylov-subspace approximation~\cite{lanczos1950iteration,al2011computing,moler2003nineteen}: (1) build a Krylov basis $V_{k_\text{iter}}$ of $H$ from $X^{(n)}_{:,k}$ by Lanczos; (2) form the $k_\text{iter}\!\times\!k_\text{iter}$ projection $T\!=\!V^{\dagger}HV$; (3) compute the small dense exponential $e^{-\mathrm{i}Tt}$; (4) return $V e^{-\mathrm{i}Tt}V^{\dagger}X^{(n)}_{:,k}$. Total cost is $\mathcal{O}(k_\text{iter}\!\cdot\!\mathrm{nnz}(H)\!\cdot\!d)$ per layer, and Appendix~\ref{proof:krylov} gives an exponential error bound that makes the approximation indistinguishable from the exact unitary at $k_\text{iter}\!=\!20$. The ``initial state'' $\psi(0)$ is simply the GNN feature column; after evolution we keep both real and imaginary parts via $f(\cdot)$.

Performing the spectral decomposition of $H$ yields:

\begin{equation}
    \label{eq:CTQW_hamiltonian}
    H = U\Lambda U^\dagger,
\end{equation}
where, $U$ is the matrix formed by the eigenvectors of $H$ and $\Lambda$ is the diagonal matrix with the corresponding eigenvalues arranged along its diagonal.
Recent studies indicate that real-world homophilic graphs consist of homophilic nodes as the majority structural pattern and heterophilic nodes in the minority~\cite{lim2021large,li2022finding}. Conversely, the situation is reversed in heterophilic graphs. However, it has been found that existing GNN model designed for homophilic graphs or heterophilic graphs only focus on one pure pattern\cite{mao2024demystifying}.   
In this context, mid-range frequency plays a more significant role compared to low and high-frequency in a mixture of homophilic and heterophilic graph~\cite{maskey2024fractional}.

According to Eq.~\ref{eq:gnn} and Eq.~\ref{eq:CTQW_hamiltonian}, We can found that CTQW-based Aggregation defines the convolutional kerne $g_{\theta}=e^{-\mathrm{i} \Lambda t}$, where $\left|\lambda_{g_{\theta}}\right|=\left|e^{-\mathrm{i} \Lambda t}\right|=1$ on every eigenmode (Appendix~\ref{proof:allpass}, Proposition~\ref{prop:allpass}).
Compared to previous approaches that designed high-pass filters for heterophilic graphs, this method is capable of aggregating not only high-frequency but also mid-range frequency. Theoretically, this would result in superior performance in real-world mixed graphs than previous methods. And  due to the norm-preserving property of CTQW, the Dirichlet energy of the graph does not decay exponentially to zero, thereby preventing over-smoothing in the CTQW-GNN model (details are provided in Section~\ref{section:oversmoothing}).

\textbf{Quantum Properties Preserved in CTQW-based Aggregation.}
A natural concern is whether CTQW-based Aggregation genuinely inherits quantum properties—\emph{superposition} and \emph{interference}—or only borrows the vocabulary. Each column $X^{(n)}_{:,k}\!=\!\sum_l \alpha_{l,k} u_l$ is a coherent linear combination over the eigenbasis $\{u_l\}$ of $H$, and the unitary propagator $e^{-\mathrm{i}Ht}=\sum_l e^{-\mathrm{i}\lambda_l t} u_l u_l^{\dagger}$ rotates \emph{all} eigenmodes by data-dependent phases while preserving their magnitudes $|\alpha_{l,k}|$. This realises (i) \emph{superposition}—features evolve as coherent combinations of eigenmodes—and (ii) \emph{interference}—distinct phases let amplitudes add coherently when projected back to the node basis, producing cancellation at near neighbours and constructive build-up at distant nodes (Figure~\ref{fig:distribution_CTQW}). The operator $f(\cdot)$ concatenates real and imaginary parts and therefore preserves phase information rather than collapsing to a modulus.

\textbf{Spectral-Gap View of Information Propagation.}
\label{para:spectralgap}
The reason CTQW-based Aggregation avoids over-smoothing while still propagating information becomes transparent through the Laplacian spectral gap. Let $0=\mu_1\le\cdots\le\mu_N$ be the eigenvalues of $L$ and denote by $\Delta:=\mu_2$ its spectral gap. Classical diffusion $e^{-Lt}$ damps the $l$-th eigenmode by $e^{-\mu_l t}$, so every non-trivial mode decays at rate $\ge e^{-\Delta t}$ and embeddings collapse onto the leading eigenvector—the origin of over-smoothing. CTQW propagation $e^{-\mathrm{i}Lt}$ instead has eigen-multipliers $e^{-\mathrm{i}\mu_l t}$ on the unit circle: \emph{no} mode is attenuated regardless of $\Delta$, mathematically guaranteeing that high- and mid-frequency content is preserved across layers. The gap instead governs the \emph{interference pattern}: larger $\Delta$ desynchronises the leading modes' phases faster, yielding the sharp ballistic peaks of Figure~\ref{fig:distribution_CTQW}.

\textbf{Walk-Time Hyperparameter $t$.}
The time parameter $t$ rescales each spectral phase $\lambda_l(H)t$; with $\lambda_{\max}=\|H\|_2$ and the above gap $\Delta$, this leads to three regimes. 
When $t\!\ll\!1/\lambda_{\max}$, $e^{-\mathrm{i}Ht}\!\approx\! I-\mathrm{i}Ht$, so propagation remains local. 
When $t\!\sim\!\pi/\Delta$, the leading modes accumulate an approximately $\pi$ phase difference, producing long range ballistic peaks that are useful for heterophilic graphs. 
When $t\!\gg\!\pi/\Delta$, the phases become densely mixed and the walk approaches its long time average. 
This analysis suggests a practical initialization range, $t\!\in\![\pi/(4\lambda_{\max}),\pi/\Delta]$, which we use across all datasets and validate in Section~\ref{sec:time_sensitivity}.

\subsection{CTQW-Attention Aggregation}
While CTQW-based aggregation captures mid-range frequencies well for mixed-pattern graphs, it can underperform on graphs that are distinctly homophilic or heterophilic. We therefore add a spatial CTQW-Attention branch that exploits CTQW-induced long-range edges to recover homophilic neighbours that are several hops away in $\mathcal{G}$. The adjacency of the new graph $\mathcal{G}_{CTQW}$ thresholds the CTQW amplitudes:
\begin{equation}
    A_{ij} =
    \begin{cases}
    1 & \left| e^{-\mathrm{i}Ht}_{ij} \right| \geq \epsilon, \\
    0 & \text{otherwise},
    \end{cases}
\end{equation}
\emph{Selecting $\epsilon$.} We use $\epsilon\!=\!5\!\times\!10^{-3}$ as the default, chosen once on the Amazon-ratings validation set and reused unchanged everywhere; a parameter-free fallback is to pick the smallest $\epsilon$ such that $\bar d(\mathcal{G}_{CTQW})\!\le\!20\bar d(\mathcal{G})$. The Lieb--Robinson bound (Appendix~\ref{proof:propagation_radius}) ensures that any $\epsilon\!\in\![10^{-3},10^{-2}]$ filters the same structurally meaningful edges (accuracy varies $<\!0.7\%$ across this range, Section~\ref{sec:epsilon_sensitivity}).

Attention from Graph Transformer aggregates information over $\mathcal{G}_{CTQW}$ only:
\begin{equation}
    X_{Attention}^{(n)}=\sum_{j \in \mathcal{N}_i}{\frac{exp(q^{T}_{i}k_{j})}{\sum_{u \in \mathcal{N}_i}{exp(q^{T}_{i}k_{u})}} v_j},
\end{equation}
where $\mathcal{N}_i$ is the set of neighbor nodes for node $i$ in the graph $\mathcal{G}_{CTQW}$. The terms Query ($q_i$), key ($k_i$) and value ($v_i$) respectively denote the outcomes derived for the feature of node $i$ $X^{(n)}_i$ in $n$-th layer of GNN after processing through a Multilayer Perceptron. $X_{Attention}^{(n)}$ is the output of n-th layer of CTQW-Attention Aggregation. With CTQW-Attention Aggregation, nodes can more effectively aggregate homophilic information from the neighbor nodes in $\mathcal{G}_{CTQW}$.
\subsection{CTQW-GNN and Over-smoothing}
\label{section:oversmoothing}
We introduces LF Aggregation for aggregating low-frequency information, thereby compensating for the shortcoming of CTQW Aggregation methods in strongly homophilic graphs. LF Aggregation can be implemented using any homophilic GNN, such as GCN, GAT, SAGE, etc. In this paper, unless specifically stated otherwise, LF Aggregation refers to the use of GAT. We define the output of the $n$-th layer LF Aggregation as $X_{LF}^{(n)}$. 

With these three aggregation methods, CTQW-GNN can aggregate mid-range frequency information through CTQW-based Aggregation,allowing it to manage mixed homophilic and heterophilic graphs, unlike most specialized GNNs. 
Additionally, CTQW-GNN utilize CTQW-Attention Aggregation to access a broader range of neighbor relationships in $\mathcal{G}_{CTQW}$ obtained through CTQW, thereby acquiring long-distance homophilic information. 
Furthermore, CTQW-GNN also aggregates low-frequency information of neighbor nodes on the original graph $\mathcal{G}$ by LF Aggregation, thus maintaining the performance of model in graphs with high homophily ratio. 

Thus, utilizing these three aggregation methods effectively addresses the issues of heterogeneous graphs; however, GNNs still face another significant challenge: Over-smoothing. Recent studies indicate that the graph attention mechanism is unable to prevent over-smoothing and results in an exponential loss of expressive capability~\cite{wu2024demystifying}. Therefore, if relying solely on LF Aggregation and CTQW-Attention Aggregation, GNNs will not be able to avoid the over-smoothing problem. CTQW-based Aggregation plays an important role in mitigating the problem of over-smoothing. Consequently, we define the input and output of each layer of the GNN as follows:
\begin{equation}
    \label{eq:feature}
    X^{(n+1)}=\sigma  \left( \left[ X^{(n)}_{LF}||X^{(n)}_{CTQW}||X^{(n)}_{Attention} \right] W_{\theta} \right)
\end{equation}
where $||$ means the concatenation along the feature dimension, $W_{\theta}$ is a learnable weight matrix. $\sigma$ is a nonlinear activation function. Based on Definition~\ref{def:oversmoothing}, we first obtain a branch-wise criterion (proved in Appendix~\ref{proof:oversmoothing_detail}):
\begin{proposition}[Branch-wise over-smoothing criterion]
    \label{proposition:oversmoothing_detail}
    Let $Y^{(n)}=[X^{(n)}_{LF}||X^{(n)}_{CTQW}||X^{(n)}_{Attention}]$ be the pre-mixing representation. $Y^{(n)}$ over-smooths if and only if the Dirichlet energies of all three branches decay exponentially to zero. For the post-mixing feature $X^{(n+1)}$, the same conclusion holds under the non-degenerate mixer condition in Appendix~\ref{proof:mixing}; in particular, any branch with non-decaying energy prevents over-smoothing whenever $W_{\theta}$ does not annihilate that branch.
\end{proposition}

In other words, the concatenated representation can collapse only when $X^{(n)}_{CTQW}$, $X^{(n)}_{Attention}$ and $X^{(n)}_{LF}$ all independently converge towards constant values. Thus, as long as one retained branch does not exponentially converge, the CTQW-GNN can avoid over-smoothing. Given this characterization, the key question is whether the CTQW branch itself has a non-decaying energy floor. We answer it with the following proposition based on the unitary property of CTQW (proof in Appendix~\ref{proof:oversmoothing_CTQW}):

\begin{proposition}[Non-decay of the CTQW branch]
    \label{proposition:oversmoothing_CTQW}
    If $H$ commutes with the Laplacian used in Dirichlet energy (e.g., $H=L$, or $H=A$ on regular graphs), $E(X^{(n)}_{CTQW})$ is invariant across layers. With the default sparse-adjacency Hamiltonian $H=A$, $E(X^{(n)}_{CTQW})$ has a positive Ces\`aro lower bound under the mild non-resonance condition stated in Appendix~\ref{proof:oversmoothing_CTQW}; therefore it cannot exponentially converge to zero.
\end{proposition}

We can observe that the CTQW-based Aggregation method plays a crucial role in mitigating the over-smoothing problem. The information aggregated through CTQW-based Aggregation is unitary rather than diffusive; it preserves non-trivial spectral content and provides a branch-level energy floor that LF or attention alone does not guarantee.

\section{Experiment}
\begin{table*}[t!]
    \centering
        \caption{Results of node classification tasks on different heterophilic datasets: mean ± std (\%). The best result ($\alpha$) for each dataset is highlight in bold and the second best ($\beta$) underlined, where Promotion is defined between these two values, i.e., $(\alpha-\beta)/\beta\times 100\%$.}
        \resizebox{\textwidth}{!}{
            \begin{tabular}{cccccccccc}
                \midrule
                {\textbf{Datasets}}

                 & \textbf{Roman-empire} & \textbf{Amazon-ratings} & \textbf{Minesweeper} & \textbf{Tolokers} & \textbf{Actor} & \textbf{Texas} & \textbf{Wiki-cooc} & \textbf{Chameleon} & \textbf{Squirrel} \\
                \textbf{Homophily Ratio} &  0.05 &  0.38 & 0.68 & 0.59 & 0.22 & 0.11 & 0.34 & 0.24 & 0.22 \\ 
                \midrule
                \midrule
                \textbf{ResNet} & $65.88{\scriptstyle\pm0.38}$ & $45.90{\scriptstyle\pm0.52}$ & $50.89{\scriptstyle\pm1.39}$ & $72.95{\scriptstyle\pm 1.06}$ & $28.75{\scriptstyle\pm0.88}$ & $80.81{\scriptstyle\pm4.75}$ & $89.36{\scriptstyle\pm0.71}$ & 
                $49.52{\scriptstyle\pm1.73} $ & $33.88{\scriptstyle\pm1.79}$ \\
                \midrule 
                \textbf{GCN} & $73.69{\scriptstyle\pm0.74}$ & $48.70{\scriptstyle\pm0.73}$ & $89.75{\scriptstyle\pm0.52}$ & $83.64{\scriptstyle\pm0.67}$ & $30.59{\scriptstyle\pm0.23}$ & $55.14{\scriptstyle\pm5.16}$ & $91.01{\scriptstyle\pm0.69}$ & 
                $50.18{\scriptstyle\pm3.29}$ & $39.06{\scriptstyle\pm1.52}$ \\
                \textbf{GAT} & $81.02{\scriptstyle\pm0.46}$ & $47.95{\scriptstyle\pm0.58}$ & $92.10{\scriptstyle\pm0.67}$ & $83.98{\scriptstyle\pm0.57}$ & $35.98{\scriptstyle\pm0.23}$ & $52.16{\scriptstyle\pm6.63}$ & $92.44{\scriptstyle\pm0.80}$ & 
                $45.02{\scriptstyle\pm1.75}$ & $32.21{\scriptstyle\pm1.63}$ \\
                \textbf{SAGE} & $85.74{\scriptstyle\pm0.67}$ & \underline{$53.63{\scriptstyle\pm0.39}$} & $93.51{\scriptstyle\pm0.57}$ & $82.43{\scriptstyle\pm0.44}$ & $36.37{\scriptstyle\pm0.21}$ & $79.03{\scriptstyle\pm1.20}$ & $93.60{\scriptstyle\pm0.31}$ & 
                $50.18{\scriptstyle\pm1.78}$ & $35.83{\scriptstyle\pm1.32}$ \\
                \midrule
                \textbf{CDE-GRAND} &  $91.64{\scriptstyle\pm0.28}$ & $47.63{\scriptstyle\pm0.43}$ & ${95.50{\scriptstyle\pm5.23}}$ & $83.98{\scriptstyle\pm0.57}$ & $34.72{\scriptstyle\pm1.40}$ & $86.22{\scriptstyle\pm3.30}$ & $97.99{\scriptstyle\pm0.38}$ & 
                $68.45{\scriptstyle\pm2.47}$ & $55.04{\scriptstyle\pm1.73}$ \\
                \textbf{GloGNN} & $59.63{\scriptstyle\pm0.69}$ & $36.89{\scriptstyle\pm0.14}$ & $51.08{\scriptstyle\pm1.23}$ & $73.39{\scriptstyle\pm1.17}$ & $37.35{\scriptstyle\pm1.30}$ & $84.32{\scriptstyle\pm4.15}$ & $88.49{\scriptstyle\pm0.45}$ & 
                $70.04{\scriptstyle\pm2.12}$ & $61.21{\scriptstyle\pm1.96}$ \\
                \textbf{PCNet} & $61.63{\scriptstyle\pm0.71}$ & $36.77{\scriptstyle\pm0.16}$ & $80.84{\scriptstyle\pm0.96}$ & $78.25{\scriptstyle\pm0.39}$ & $37.02{\scriptstyle\pm0.87}$ & $88.11{\scriptstyle\pm2.17}$ & $ 87.35{\scriptstyle\pm0.29}$ & 
                $73.55{\scriptstyle\pm1.26}$ & $63.53{\scriptstyle\pm0.26}$ \\
                \textbf{EG-GCN} & $63.32{\scriptstyle\pm0.32}$ & $38.33{\scriptstyle\pm0.78}$ & $79.67{\scriptstyle\pm0.89}$ & $81.41{\scriptstyle\pm1.16}$ & $37.80{\scriptstyle\pm0.60}$ & $88.92{\scriptstyle\pm3.30}$ & $85.56{\scriptstyle\pm0.83}$ & 
                $71.93{\scriptstyle\pm2.58}$ & $61.24{\scriptstyle\pm1.97}$ \\
                \midrule
                \textbf{GPRGNN} & $64.85{\scriptstyle\pm0.27}$ & $44.88{\scriptstyle\pm0.34}$ & $86.24{\scriptstyle\pm0.61}$ & $72.94{\scriptstyle\pm0.97}$ & $34.63{\scriptstyle\pm1.22}$ & $78.38{\scriptstyle\pm4.36}$ & $91.90{\scriptstyle\pm0.78}$ & 
                $47.26{\scriptstyle\pm1.74}$ & $33.39{\scriptstyle\pm2.05}$ \\
                \textbf{FSGNN} & $79.92{\scriptstyle\pm0.56}$ &  $52.74{\scriptstyle\pm0.83}$ & $90.08{\scriptstyle\pm0.70}$ & $82.76{\scriptstyle\pm0.61}$ & $35.38{\scriptstyle\pm0.81}$ & $87.57{\scriptstyle\pm4.71}$ & $91.83{\scriptstyle\pm0.53}$ & 
                \underline{$77.85{\scriptstyle\pm0.46}$} & \underline{$68.93{\scriptstyle\pm1.69}$} \\
                \textbf{ACMP-GCN} & $71.27{\scriptstyle\pm0.59}$ & $44.76{\scriptstyle\pm0.52}$ & $76.15{\scriptstyle\pm1.12}$ & $75.03{\scriptstyle\pm0.92}$ & $37.57{\scriptstyle\pm0.31}$ & $86.23{\scriptstyle\pm3.05}$ & $92.68{\scriptstyle\pm0.37}$ & 
                $69.04{\scriptstyle\pm1.74}$ & $58.02{\scriptstyle\pm1.86}$  \\
                \textbf{FLODE} & $74.97{\scriptstyle\pm0.53}$ & $48.75{\scriptstyle\pm0.82}$ & $92.43{\scriptstyle\pm0.51}$ & $84.17{\scriptstyle\pm0.58}$ & $37.16{\scriptstyle\pm1.42}$ & $71.08{\scriptstyle\pm0.72}$ & $92.51{\scriptstyle\pm0.97}$ & $73.60{\scriptstyle\pm1.55}$ & $64.23{\scriptstyle\pm1.84}$ \\
                \textbf{UniFilter} & $63.45{\scriptstyle\pm0.23}$ & $43.42{\scriptstyle\pm0.65}$ & $76.58{\scriptstyle\pm2.13}$ & $78.26{\scriptstyle\pm0.67}$ & $37.79{\scriptstyle\pm1.11}$ & $81.12{\scriptstyle\pm1.63}$ & $83.85{\scriptstyle\pm0.65}$ & 
                $73.66{\scriptstyle\pm2.44}$ & $64.26{\scriptstyle\pm1.46}$ \\    

                \textbf{F-GRAND} & $93.06{\scriptstyle\pm0.55}$ & $49.01{\scriptstyle\pm0.56}$ & $96.04{\scriptstyle\pm0.25}$ & $82.68{\scriptstyle\pm0.86}$ & $38.28{\scriptstyle\pm0.74}$ & $92.97{\scriptstyle\pm4.39}$ & $98.73{\scriptstyle\pm0.68}$ &
                $71.45{\scriptstyle\pm1.98}$ & $60.86{\scriptstyle\pm1.05}$ \\

                \textbf{D-CDE} & $93.87{\scriptstyle\pm0.41}$ & $49.43{\scriptstyle\pm1.26}$ & $96.47{\scriptstyle\pm1.89}$ & $83.02{\scriptstyle\pm0.86}$ & $38.37{\scriptstyle\pm1.55}$ & $91.83{\scriptstyle\pm5.47}$ & $98.58{\scriptstyle\pm0.12}$ &
                $64.88{\scriptstyle\pm2.18}$ & $45.62{\scriptstyle\pm2.12}$ \\
                \textbf{CGNN} & $92.16{\scriptstyle\pm0.46}$ & $50.32{\scriptstyle\pm0.41}$ & $95.18{\scriptstyle\pm0.55}$ & $83.65{\scriptstyle\pm0.46}$ & $37.92{\scriptstyle\pm1.27}$ & $90.27{\scriptstyle\pm3.62}$ & $98.21{\scriptstyle\pm0.51}$ & $75.64{\scriptstyle\pm1.31}$ & $66.18{\scriptstyle\pm1.45}$ \\
                \textbf{AMP} & $92.45{\scriptstyle\pm0.55}$ & $51.18{\scriptstyle\pm0.58}$ & $95.89{\scriptstyle\pm0.43}$ & $83.46{\scriptstyle\pm0.41}$ & $37.90{\scriptstyle\pm0.93}$ & $89.46{\scriptstyle\pm4.31}$ & $98.30{\scriptstyle\pm0.27}$ & $73.84{\scriptstyle\pm1.95}$ & $63.21{\scriptstyle\pm1.78}$ \\
                \textbf{HeroFilter} & \underline{$94.02{\scriptstyle\pm0.36}$} & $52.13{\scriptstyle\pm0.41}$ & \underline{$96.51{\scriptstyle\pm0.31}$} & \underline{$84.30{\scriptstyle\pm0.39}$} & \underline{$38.59{\scriptstyle\pm1.09}$} & \underline{$93.51{\scriptstyle\pm3.10}$} & \underline{$98.81{\scriptstyle\pm0.41}$} & $76.78{\scriptstyle\pm1.65}$ & $67.45{\scriptstyle\pm1.41}$ \\
                \midrule
                \textbf{CTQW-GNN (ours)} &  \bm{$94.13{\scriptstyle\pm0.42}$} & \bm{$54.07{\scriptstyle\pm0.63}$} & \bm{$97.32{\scriptstyle\pm0.34}$} & \bm{$85.22{\scriptstyle\pm0.43}$} & \bm{$38.92{\scriptstyle\pm2.55}$} &  \bm{$93.93{\scriptstyle\pm2.78}$} & \bm{$99.65{\scriptstyle\pm0.22}$} &
                \bm{$79.23{\scriptstyle\pm1.63}$} & \bm{$70.82{\scriptstyle\pm1.31}$} \\

                \textbf{Promotion $\uparrow$ (\%)} & {$0.12$} & {$0.82$} & {$0.84$} & {$1.09$} & {$0.85$} &  {$0.45$} & {$0.85$} &
                1.77 & 2.74 \\
                \midrule
            \end{tabular}
        }
    \vspace{0pt}

    \label{tab:main_result}
\end{table*}

\begin{table}[t!]
    \centering
        \caption{Results of node classification tasks on different homophilic datasets: mean ± std (\%).}
        \resizebox{0.5\textwidth}{!}{
            \begin{tabular}{cccccc}
                \midrule
                {\textbf{Datasets}}

                 & \textbf{Cora} & \textbf{Citeseer} & \textbf{PubMed} & \textbf{Computer} & \textbf{Photo} \\
                \textbf{Homophily Ratio} & 0.81 & 0.72 & 0.79 & 0.78 & 0.83 \\ 
                \midrule
                \midrule
                \textbf{ResNet} & $76.44{\scriptstyle\pm0.30}$ & $76.25{\scriptstyle\pm0.28}$ & $86.43{\scriptstyle\pm0.13}$ & 
                $84.68{\scriptstyle\pm0.78}$ & $91.49{\scriptstyle\pm0.30}$ \\
                \midrule 
                \textbf{GCN} & $87.78{\scriptstyle\pm0.96}$ & $81.39{\scriptstyle\pm1.23}$ & $88.90±{\scriptstyle\pm0.32}$ & 
                $83.55{\scriptstyle\pm0.38}$ & $89.30{\scriptstyle\pm0.44}$ \\
                \textbf{GAT} & $76.70{\scriptstyle\pm0.42}$ & $67.20{\scriptstyle\pm0.46}$ & $83.28{\scriptstyle\pm0.12}$ & 
                $85.36{\scriptstyle\pm0.50}$ & $90.81{\scriptstyle\pm0.22}$ \\
                \textbf{SAGE} & $86.58{\scriptstyle\pm0.26}$ & $78.24{\scriptstyle\pm0.30}$ & $86.85{\scriptstyle\pm0.11}$ & $83.11{\scriptstyle\pm0.23}$ & $90.51{\scriptstyle\pm0.25}$ \\
                \midrule
                \textbf{CDE-GRAND}& $87.19{\scriptstyle\pm1.44}$ & $80.04{\scriptstyle\pm1.75}$ & $90.05{\scriptstyle\pm0.64}$ & 
                $81.95{\scriptstyle\pm0.43}$ & $88.27{\scriptstyle\pm1.94}$ \\
                \textbf{GloGNN} & $88.31{\scriptstyle\pm1.15}$ & $77.41{\scriptstyle\pm1.65}$ & $89.62{\scriptstyle\pm0.35}$  & 
                $87.69{\scriptstyle\pm0.33}$ & $92.81{\scriptstyle\pm0.13}$ \\
                \textbf{PCNet} & $82.81{\scriptstyle\pm0.50}$ & $69.92{\scriptstyle\pm0.70}$ & $80.01{\scriptstyle\pm0.88}$ & 
                $83.29{\scriptstyle\pm1.53}$ & $89.51{\scriptstyle\pm1.04}$ \\
                \textbf{EG-GCN} & $88.07{\scriptstyle\pm1.16}$ & $78.09{\scriptstyle\pm1.20}$ & $89.64{\scriptstyle\pm0.38}$ & $85.67{\scriptstyle\pm2.32}$ & $85.33{\scriptstyle\pm1.68}$ \\
                \midrule
                \textbf{GPRGNN} & $79.51{\scriptstyle\pm0.36}$ & $67.63{\scriptstyle\pm0.38}$ & $85.07{\scriptstyle\pm0.09}$ & 
                $87.63{\scriptstyle\pm0.48}$ & $94.60{\scriptstyle\pm0.30}$ \\
                \textbf{FSGNN} & $87.73{\scriptstyle\pm1.34}$ & $77.19{\scriptstyle\pm0.81}$ & $89.73{\scriptstyle\pm3.16}$ & 
                \underline{$91.53{\scriptstyle\pm1.03}$} & \underline{$95.07{\scriptstyle\pm0.53}$} \\
                \textbf{ACMP-GCN} & $84.87{\scriptstyle\pm0.67}$ & $75.96{\scriptstyle\pm1.07}$ & $78.93{\scriptstyle\pm1.01}$ & 
                $83.56{\scriptstyle\pm1.42}$ & $91.83{\scriptstyle\pm1.17}$ \\
                \textbf{FLODE} & $86.44{\scriptstyle\pm1.17}$ & $78.07{\scriptstyle\pm1.62}$ & $89.02{\scriptstyle\pm0.38}$ & $88.44{\scriptstyle\pm0.43}$ & $93.86{\scriptstyle\pm0.85}$ \\
                \textbf{UniFilter} & \underline{$89.12{\scriptstyle\pm0.87}$} &  $80.28{\scriptstyle\pm1.31}$ & $90.19{\scriptstyle\pm0.41}$ &
                $87.98{\scriptstyle\pm0.56}$ & $94.03{\scriptstyle\pm0.78}$ \\

                \textbf{F-GRAND} & $83.15{\scriptstyle\pm1.19}$ & $73.96{\scriptstyle\pm1.52}$  & $79.38{\scriptstyle\pm1.55}$ &
                $84.40{\scriptstyle\pm1.50}$ & $92.80{\scriptstyle\pm0.60}$ \\

                \textbf{D-CDE} &  $84.17{\scriptstyle\pm1.22}$ & $74.03{\scriptstyle\pm2.13}$ &  $79.52{\scriptstyle\pm1.18}$ &
                $87.30{\scriptstyle\pm1.30}$ & $94.10{\scriptstyle\pm0.70}$ \\
                \midrule
                \textbf{CGNN} & $87.62{\scriptstyle\pm1.34}$ & $78.93{\scriptstyle\pm1.50}$ & $89.51{\scriptstyle\pm0.45}$ & $88.42{\scriptstyle\pm0.61}$ & $93.84{\scriptstyle\pm0.45}$ \\
                \textbf{AMP} & $88.30{\scriptstyle\pm0.92}$ & $79.71{\scriptstyle\pm1.28}$ & $89.86{\scriptstyle\pm0.38}$ & $89.04{\scriptstyle\pm0.53}$ & $94.21{\scriptstyle\pm0.40}$ \\
                \textbf{HeroFilter} & $88.85{\scriptstyle\pm1.05}$ & \underline{$80.31{\scriptstyle\pm1.18}$} & \underline{$90.42{\scriptstyle\pm0.36}$} & $90.71{\scriptstyle\pm0.42}$ & $94.86{\scriptstyle\pm0.31}$ \\
                \midrule
                \textbf{CTQW-GNN (ours)} & \bm{$91.31{\scriptstyle\pm2.12}$} & \bm{$82.97{\scriptstyle\pm1.37}$} & \bm{$91.38{\scriptstyle\pm0.79}$} &
                \bm{$92.48{\scriptstyle\pm0.13}$} &  \bm{$95.87{\scriptstyle\pm0.35}$} \\

                \textbf{Promotion $\uparrow$ (\%)} & 2.46 & 3.31 & 1.06 &
                $1.04$ & $0.84$ \\ 
                \midrule
            \end{tabular}
        }
    \vspace{0pt}

    \label{tab:main_result_homo}
    
    \vspace{-15pt}
\end{table}

We validate CTQW-GNN on homophilic and heterophilic node classification, then probe node subgroups, ablations, and hyperparameter sensitivity.
\subsection{Experimental Setup}
{\textbf{Datasets and Metrics. }}
We use five homophilic datasets (Cora~\cite{mccallum2000automating}, Citeseer~\cite{sen2008collective}, PubMed~\cite{namata2012query}, Computers, Photo~\cite{namata2012query}) and nine heterophilic datasets: the classical four (Chameleon, Squirrel, Actor, Texas)~\cite{pei2020geom} and the five new datasets of~\citet{platonov2023critical} (Roman-empire, Amazon-ratings, Minesweeper, Tolokers, Wiki-cooc). Splits follow~\cite{zheng2023finding} for homophilic data, \cite{yan2022two} for the classical heterophilic four, and the original papers otherwise. We report ROC-AUC for the binary-class Minesweeper/Tolokers and accuracy elsewhere.

{\textbf{Baselines and Implementation. }}We compare with five categories: graph-agnostic ResNet~\cite{he2016deep}; classical GNNs (GCN~\cite{kipf2017semi}, GAT~\cite{velickovic2018graph}, SAGE~\cite{hamilton2017inductive}); heterophily-specific (CDE-GRAND~\cite{zhao2023graph}, GloGNN~\cite{li2022finding}, EG-GCN~\cite{liu2025integrating}, PCNet~\cite{li2024pc}); both-problem-aware (GPR-GNN~\cite{chien2021adaptive}, FSGNN~\cite{maurya2022simplifying}, ACMP-GCN~\cite{wang2023acmp}, FLODE~\cite{maskey2024fractional}, F-GRAND~\cite{kangunleashing}, D-CDE~\cite{zhaodistributed}, UniFilter~\cite{huang2024universal}, CGNN~\cite{zhuo2025commute}, AMP~\cite{errica2025adaptive}, and HeroFilter~\cite{zhang2025herofilter}). CTQW-GNN uses the adjacency matrix as $H$ and GAT as the LF Aggregation.

\subsection{Main Experimental Results}
\begin{figure}[b!]    
    \centering
    \begin{subfigure}[b]{0.23\textwidth}
        \centering
        \includegraphics[width=\linewidth]{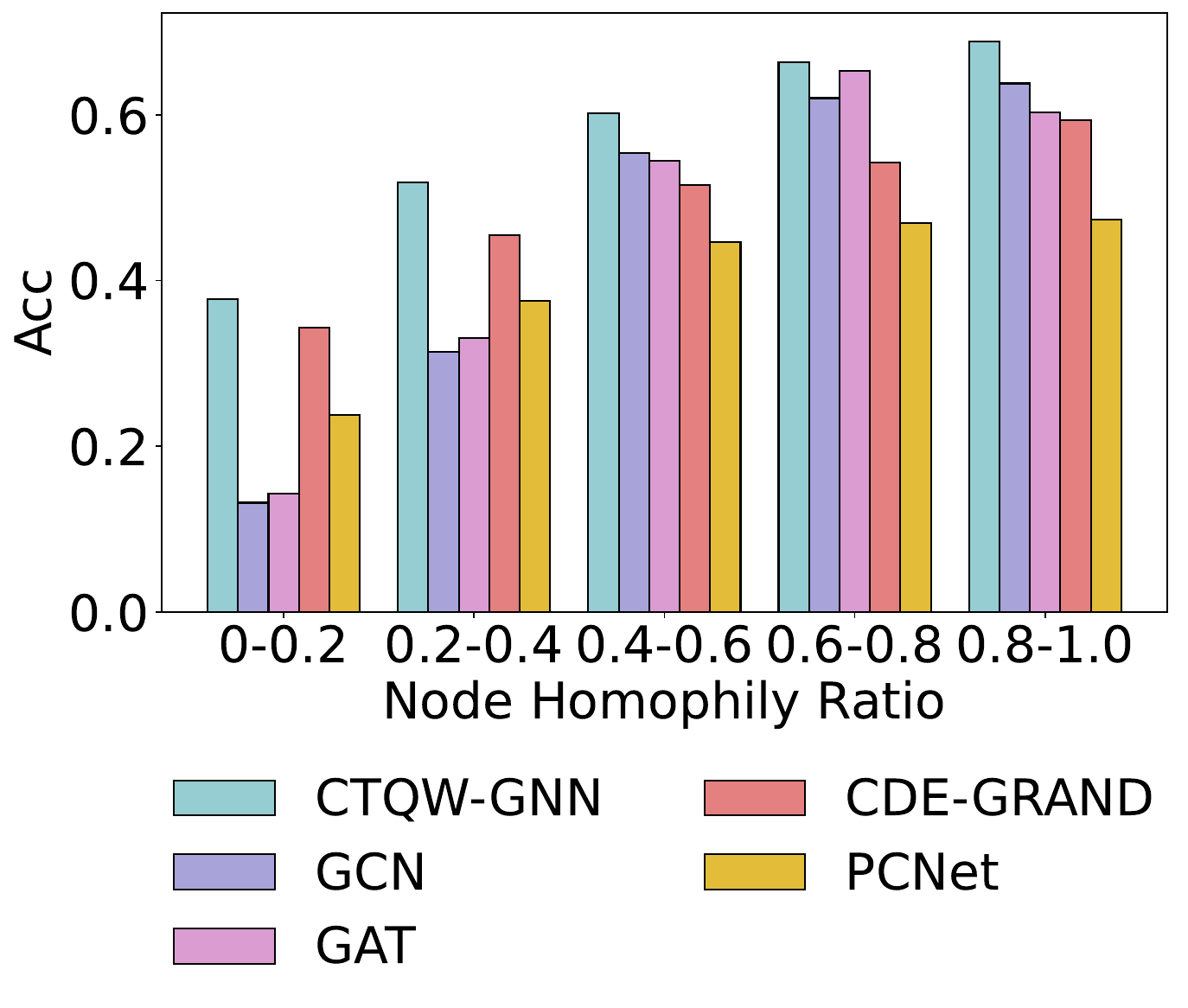}
        \caption{Results on Amazon-ratings.}
        \label{fig:amazon}
    \end{subfigure}
    \hfill
    \begin{subfigure}[b]{0.23\textwidth}
        \centering
        \includegraphics[width=\textwidth]{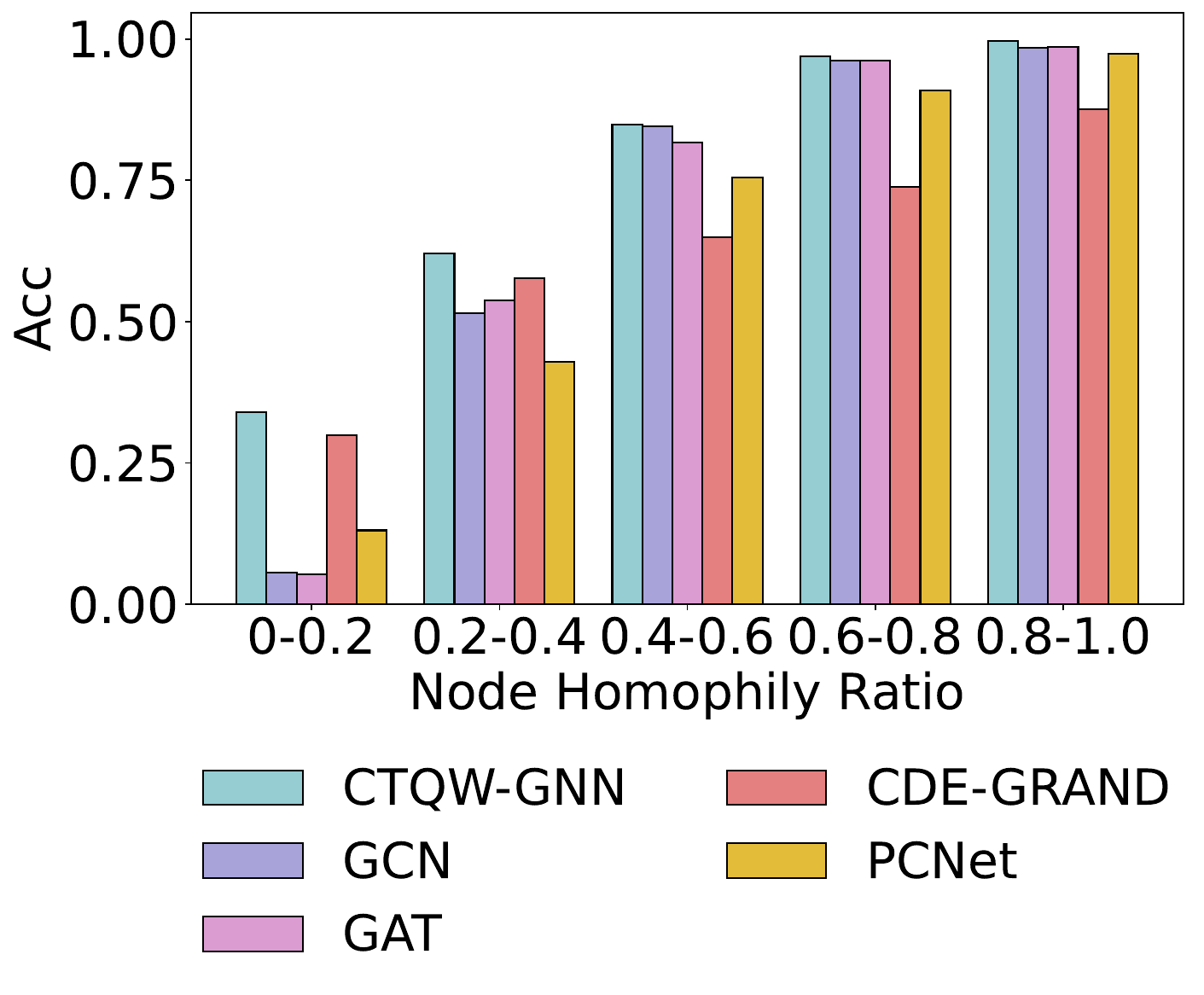}
        \caption{Results on Computer.}
        \label{fig:computer}
    \end{subfigure}
    \vspace{-5pt}
    \caption{Performance on node subgroups.}
    \label{fig:subgroup}
    \vspace{-15pt}
\end{figure}

We conduct extensive experiments in homophilic
and heterophilic graph. We present the node classification results on heterophilic graphs in Table~\ref{tab:main_result}, while the results on homophilic graphs are reported in Table~\ref{tab:main_result_homo}. 

Firstly, it is evident that CTQW-GNN model consistently performs best across all homophily ratio. This advantage arises from our proposed information aggregation methods. In cases of low homophily ratio, CTQW-Attention Aggregation captures long-range dependencies and CTQW-based Aggregation accesses high-frequency information, enhancing GNN performance on heterophilic graphs. For highly homophilic scenarios, LF Aggregation gathers low-frequency information, improving GNN performance on homophilic graphs. When the graph presents a mix of homophilic and heterophilic partern, mid-frequency information from CTQW-based Aggregation becomes critical.

\begin{figure*}[ht!]
    
    \centering
    \small
    \includegraphics[width=\textwidth]{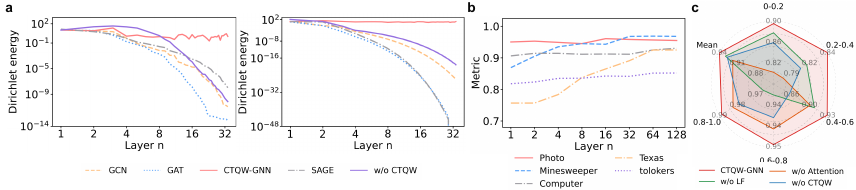}    
    \vspace{-20pt}
    \caption{\textbf{(a).} Dirichlet energy of graph. Left: Minesweeper results; Right: Amazon-ratings results. \textbf{(b.)} Accuracy of CTQW-GNN. Figure~\textbf{a} and \textbf{b} are the performance of GNN models with different depth. In Figure~\textbf{b}, the metric is ROC-AUC for the Minesweeper dataset and accuracy for the other datasets. "w/o" represents the removal of the corresponding aggregation method. \textbf{(c).} Impact of information aggregation methods. The axes denote the accuracy of models on Minesweeper dataset.}
    \label{fig:DE_ACC_radar}
\end{figure*}
Secondly, In the new dataset introduced by~\cite{platonov2023critical}, most heterophily-specific GNNs significantly underperform compared to classic homophilic GNNs (\eg SAGE achieves an average rank of 4.4, placing it second). Some methods even perform worse than graph-agnostic ResNet. This underscores the significant role of this new dataset in validating the effectiveness of heterophilic GNNs. Under these conditions, CTQW still outperforms most baseline models (outperforming the nearest rival by 1.48\% on average). This outcome further demonstrates the efficacy of CTQW-GNN in tackling the critical challenge of heterophilic. 

Finally, it is evident that most GNNs designed to address both over-smoothing and heterophily problem underperform compared to those heterophily-specific models. This observation suggests that these methods sacrifice performance to mitigate over-smoothing. However, CTQW-GNN demonstrates superior results across all datasets, indicating a successful resolution of this challenge. CTQW-GNN not only effectively mitigates over-smoothing problem (Section~\ref{section:oversmoothing}) but also improves GNN efficacy on heterophilic graphs.
\subsection{Further Analysis}
{\textbf{Performance on Node Subgroups. }}
Recent studies criticize existing methods for focusing solely on purely homophilic or heterophilic graphs, ignoring the fact that real-world graphs typically exhibit a mixture of homophilic and heterophilic patterns~\cite{mao2024demystifying}. This means that homophilic graphs may contain heterophilic nodes, and vice versa; these nodes can often become bottlenecks that limit GNN performance. To further validate the superiority of CTQW-GNN in handling these mixed graphs, we divided the nodes of Computers and Amazon-ratings into Node Subgroups based on node homophily ratio, and then tested the performance of various GNN models on these different Node Subgroups. The results are shown in Figure~\ref{fig:subgroup}.

On heterophilic nodes, heterophily-specific GNNs outperform classic GNNs and vice versa for homophilic nodes; nodes that deviate from a graph's typical pattern are the dominant performance bottleneck. CTQW-GNN outperforms existing methods on \emph{any} node homophily ratio in both regimes, thanks to its simultaneous aggregation of low-, mid-, and high-frequency neighbor information together with long-range dependencies.

{\textbf{Evolution of Dirichlet Energy. }}
The norm-preserving property of CTQW prevents node features from collapsing to a constant. We track the Dirichlet energy on Minesweeper and Amazon-ratings (Figure~\ref{fig:DE_ACC_radar}a).

Figure~\ref{fig:DE_ACC_radar}a demonstrates that the Dirichlet energy calculated from layer-wise node features via classic GNN methods (\ie~GCN, GAT) converges exponentially with the increasing layers. In contrast, the Dirichlet energy in CTQW  consistently stabilizes around a constant value. Notably, when the CTQW-based Aggregation is not utilized (\ie~w/o CTQW), the Dirichlet energy of the graph initially remains stable for a period; however, it still rapidly converges exponentially. This further substantiates the efficacy of CTQW-based Aggregation in mitigating the over-smoothing problem.

{\textbf{Ablation Study}}
Our approach consists of three information aggregation methods: (1) CTQW-based Aggregation, (2) CTQW-Attention Aggregation, (3) LF Aggregation. To verify the function of each module and its efficacy in enhancing model performance, we conducted ablation experiments on the Minesweeper dataset to analyze the impact of these aggregation methods on GNN performance. We consider removing the three aggregation, respectively, and the ablation results on node Subgroups are shown in Figure~\ref{fig:DE_ACC_radar}c.


Firstly, it is evident that the removal of any aggregation method results in a reduction of the radar chart area, indicating that all three information aggregation methods significantly enhance model performance. Secondly, the removal of LF, CTQW-based, and CTQW-Attention aggregations leads to different degrees of performance degradation across GNN node subgroups. Specifically, omitting CTQW-Attention hurts low-homophily subgroups ($0$--$0.4$) the most because long-range homophilic retrieval is crucial there, while omitting LF hurts high-homophily subgroups ($0.6$--$1.0$) where low-pass aggregation is beneficial. Removing CTQW-based aggregation causes the largest drop on mixed-pattern nodes ($0.4$--$0.6$), where mid-frequency information is most important. Furthermore, as Figure~\ref{fig:DE_ACC_radar}a illustrates, removing CTQW-based aggregation also reintroduces over-smoothing.

{\textbf{Sensitivity to Walk Time $t$.}}
\label{sec:time_sensitivity}
Walk time $t$ controls the propagation range of the CTQW branch. We evaluate its influence by sweeping nine values from $10^{-6}$ to $10^{2}$ on four representative datasets, as shown in Table~\ref{tab:walk_time}. When $t$ is extremely small ($10^{-6}$--$10^{-4}$), the CTQW propagation is weak and the performance is relatively low. As $t$ increases to the middle range ($10^{-3}$--$10^{-1}$), the performance improves substantially and remains close to the best result on all four datasets. For example, the results on Roman-empire, Amazon-ratings, and Minesweeper all stay within $1\%$ of their best scores across this range, while Cora also keeps a stable plateau from $10^{-2}$ to $1$.
The results indicate that CTQW-GNN is not sensitive to a precise choice of $t$. The best performance does not appear at a single isolated value; instead, each dataset has a broad stable region where different $t$ values lead to very similar accuracy. Performance only drops clearly when $t$ is too small or too large. Therefore, CTQW-GNN shows robust performance with respect to the walk time hyperparameter.

\begin{table}[t!]
    \centering
    \scriptsize
    \setlength{\tabcolsep}{2pt}
    \caption{Pserformance of CTQW-GNN under different walk time $t$.}
    \label{tab:walk_time}
    \resizebox{\columnwidth}{!}{
    \begin{tabular}{l|ccccccccc}
    \toprule
    $t$ & $10^{-6}$ & $10^{-5}$ & $10^{-4}$ & $10^{-3}$ & $10^{-2}$ & $10^{-1}$ & $1$ & $10$ & $10^{2}$ \\
    \midrule
    Cora          & 86.74 & 87.05 & 87.62 & 88.93 & 90.27 & \textbf{91.31} & 91.18 & 88.52 & 84.81 \\
    Roman-empire  & 84.55 & 84.92 & 85.78 & 93.34 & \textbf{94.13} & 93.88 & 92.20 & 88.46 & 85.59 \\
    Amazon-ratings& 50.71 & 51.04 & 51.92 & 53.59 & \textbf{54.07} & 53.90 & 53.21 & 50.93 & 48.66 \\
    Minesweeper   & 87.79 & 88.05 & 89.20 & 96.80 & \textbf{97.32} & 97.07 & 94.93 & 92.30 & 89.63 \\
    \bottomrule
    \end{tabular}
    }
    \vspace{-6pt}
\end{table}

\noindent\textbf{Sensitivity to the Sparsification Threshold $\epsilon$.}
\label{sec:epsilon_sensitivity}
The threshold $\epsilon$ controls how many CTQW edges are kept: a smaller $\epsilon$ keeps more long-range connections, while a larger $\epsilon$ prunes more connections. We test five values of $\epsilon$ on four representative datasets, and the results are shown in Table~\ref{tab:epsilon}.

Overall, CTQW-GNN is robust to $\epsilon$. When $\epsilon$ changes from $10^{-4}$ to $10^{-2}$, the performance on all four datasets remains very stable and close to the best result. This shows that the model does not rely on one specific sparsification threshold. When $\epsilon$ is too small, many weak CTQW connections are retained, which brings more computation but gives little or no accuracy improvement. When $\epsilon$ is too large, too many useful long-range connections are removed, and the performance drops clearly. Therefore, a moderate threshold such as $5\!\times\!10^{-3}$ provides a good balance between performance and sparsity.

\begin{table}[t!]
    \centering
    \scriptsize
    \setlength{\tabcolsep}{3pt}
    \caption{Effect of sparsification threshold $\epsilon$.}
    \label{tab:epsilon}
    \begin{tabular}{l|c|c|c|c|c}
    \toprule
    Dataset & $10^{-4}$ & $10^{-3}$ & $5\!\times\!10^{-3}$ & $10^{-2}$ & $5\!\times\!10^{-2}$ \\
    \midrule
    Amazon-ratings & 53.92 & 54.03 & \textbf{54.07} & 53.74 & 51.18 \\
    Minesweeper    & 97.28 & 97.21 & \textbf{97.32} & 96.95 & 94.40 \\
    Roman-empire   & 94.09 & 94.03 & \textbf{94.13} & 93.71 & 90.96 \\
    Wiki-cooc      & 99.56 & 99.58 & \textbf{99.65} & 99.22 & 96.48 \\
    \bottomrule
    \end{tabular}
\end{table}

\subsection{Time Complexity Analysis}
\label{sec:complexity}
Although $e^{-\mathrm{i}Ht}$ appears dense, CTQW-GNN never forms this operator explicitly. The graph $\mathcal{G}_{CTQW}$ is built once via a sparse degree-$K$ Chebyshev expansion ($K\!=\!20$) at cost $\mathcal{O}(K|\mathcal{E}|)$, amortised over all epochs. Per layer, the CTQW branch applies $e^{-\mathrm{i}Ht}X^{(n)}$ via a $k$-step Krylov projection ($k\!=\!20$) at $\mathcal{O}(k|\mathcal{E}|d)$; the LF and CTQW-Attention branches each cost $\mathcal{O}(|\mathcal{E}|d+Nd^{2})$, the same as GCN~\cite{kipf2017semi} and GAT~\cite{velickovic2018graph}. Since the sparsification rule (Section~\ref{sec:epsilon_sensitivity}) keeps $|\mathcal{E}_{CTQW}|\!=\!\mathcal{O}(|\mathcal{E}|)$, the total per-layer cost is $\mathcal{O}\!\big((k{+}1)|\mathcal{E}|d+Nd^{2}\big)$—linear in $|\mathcal{E}|$ and matching GCN/GAT up to a small constant factor $k{+}1\!\le\!21$.

\section{Conclusion}
Existing GNN methods predominantly address over-smoothing and heterophily separately, or mitigate over-smoothing at the cost of expressiveness. We propose CTQW-GNN, integrating three aggregators—two CTQW-inspired and one classical—to tackle both issues by aggregating low-, mid-, high-frequency, and long-range information through a unitary, norm-preserving propagator. Theoretically, the CTQW branch has a positive spectral/Ces\`aro energy floor and therefore cannot decay exponentially (Prop.~\ref{proposition:oversmoothing_CTQW}, Lemma~\ref{lemma:spectralgapbound}); empirically, CTQW-GNN reaches state-of-the-art accuracy on all 14 benchmarks at a per-edge cost asymptotically equal to GCN/GAT. Future work will study learnable low-pass companions and graph-Hamiltonian design.
\section{Acknowledgments}
The work is supported by the National Natural Science Foundation of China (No. 62276269 and No. 62576331), which is greatly appreciated. 


\setcounter{secnumdepth}{2}

\setcounter{figure}{0}
\setcounter{table}{0}
\setcounter{equation}{0}
\setcounter{algorithm}{0}
\setcounter{definition}{0}

\renewcommand{\theequation}{S.\arabic{equation}}
\renewcommand{\thefigure}{S.\arabic{figure}}
\renewcommand{\thetable}{S.\arabic{table}}
\renewcommand{\thealgorithm}{S.\arabic{algorithm}}
\appendix
\vspace{6pt}

\section{Proof}
\label{proof:notation}
\paragraph{Notation.}
$\mathcal{G}=(\mathcal{V},\mathcal{E})$, $N\!=\!|\mathcal{V}|$, $L\!=\!I\!-\!D^{-1/2}AD^{-1/2}$. The Hamiltonian $H$ is Hermitian (default $H\!=\!A$). $L\!=\!V\Lambda_L V^{\!\dagger}$ with eigenvalues $0\!=\!\mu_1\!\le\!\mu_2\!\le\!\cdots\!\le\!\mu_N$ and gap $\Delta\!:=\!\mu_2$. Let $P_{\perp}\!=\!I-v_1v_1^{\dagger}$ be the projection onto the non-constant Laplacian subspace. For $X\!\in\!\mathbb{C}^{N\times d}$ define spectral coordinates $\tilde X\!:=\!V^{\!\dagger}X$ (rows indexed by mode $\ell$). The matrix-form Dirichlet energy
\begin{equation}\label{eq:matrixform}
E(X)\;=\;\mathrm{tr}(X^{\!\dagger}LX)\;=\;\sum_{\ell=1}^{N}\mu_\ell\,\|\tilde X_\ell\|_2^2
\end{equation}
agrees with Eq.~\ref{eq:DE} up to a constant fixed by $\mathcal{G}$~\cite{chung1997spectral}. Standing assumptions (verified on every benchmark): \textbf{(A1)} $\Delta\!>\!0$; \textbf{(A2)} $\|P_{\perp}X^{(0)}\|_F\!>\!0$. For the default $H\!=\!A$ we additionally use the standard non-resonance condition \textbf{(A3)} that the CTQW orbit does not concentrate all feature energy in the constant Laplacian mode; equivalently, $\liminf_{T\to\infty}T^{-1}\sum_{n<T}\|P_{\perp}e^{-\mathrm{i}nHt}X^{(0)}\|_F^2>0$. This excludes a measure-zero set of walk times and initial features, and is directly checked by the branch-energy diagnostic in Figure~\ref{fig:DE_ACC_radar}a.

\subsection{Proof of Proposition~\ref{proposition:oversmoothing_detail}}
\label{proof:oversmoothing_detail}
\begin{proof}
Let $Y^{(n)}\!=\![X^{(n)}_{LF}\!\,\Vert\,\!X^{(n)}_{CTQW}\!\,\Vert\,\!X^{(n)}_{Att}]\!\in\!\mathbb{C}^{N\times 3d}$ be the pre-mixing concatenation. By Eq.~\ref{eq:matrixform} and the column-block structure of $Y^{(n)}$,
\begin{equation}\label{eq:Esum}
E(Y^{(n)})=\mathrm{tr}(Y^{(n)\dagger}L Y^{(n)})=E_{LF}^{(n)}+E_{CTQW}^{(n)}+E_{Att}^{(n)},
\end{equation}
since $\mathrm{tr}([P\,\Vert\,Q]^{\dagger}L[P\,\Vert\,Q])=\mathrm{tr}(P^{\!\dagger}LP)+\mathrm{tr}(Q^{\!\dagger}LQ)$ for any pair of matrices $P,Q$.
Because all three terms are non-negative, $E(Y^{(n)})\le ae^{-bn}$ holds iff each branch energy is bounded by an exponential envelope (with rate at least $b$ in the ``only-if'' direction and the minimum branch rate in the ``if'' direction). This proves the exact branch-wise criterion for the representation before the learned mixer. For the post-mixing features $X^{(n+1)}\!=\!\sigma(Y^{(n)}W_\theta)$, Lemma~\ref{lemma:concat} gives the always-valid upper bound, and Lemma~\ref{lemma:mixing} gives the corresponding lower bound whenever the learned projection keeps a non-zero CTQW component. Thus the criterion transfers to the actual layer output under the non-degenerate mixer condition stated in Proposition~\ref{proposition:oversmoothing_detail}.
\end{proof}

\begin{lemma_app}\label{lemma:concat}
Let $Y\!\in\!\mathbb{C}^{N\times m}$, $W\!\in\!\mathbb{R}^{m\times d}$, and $\sigma$ be entrywise $\rho_+$-Lipschitz with $\sigma(0)\!=\!0$. Then $E(\sigma(YW))\!\le\!\rho_+^2\|W\|_2^2 E(Y)$. If, on the subspace spanned by layer-wise edge differences, the map $z\mapsto\sigma(zW)$ has gain at least $\rho_->0$, then $E(\sigma(YW))\!\ge\!\rho_-^2 E(Y)$ on that subspace.
\end{lemma_app}
\begin{proof}
Write $E(Z)=\sum_{(i,j)\in\mathcal{E}}\|z_i/\!\sqrt{d_i}-z_j/\!\sqrt{d_j}\|_2^2$. For each edge term, entrywise Lipschitzness gives $\|\sigma(u)\!-\!\sigma(v)\|_2\!\le\!\rho_+\|u\!-\!v\|_2$, and $\|(u-v)^{\top}W\|_2\!\le\!\|W\|_2\|u-v\|_2$ yields the upper bound after summing over edges. The lower bound is exactly the stated non-degenerate gain condition applied to the same edge-difference terms. Degenerate projections that cancel an entire branch are therefore excluded explicitly rather than hidden inside an invalid full-rank assumption.
\end{proof}

\subsection{Proof of Proposition~\ref{proposition:oversmoothing_CTQW}}
\label{proof:oversmoothing_CTQW}
\begin{proof}
Let $U_t\!:=\!e^{-\mathrm{i}Ht}$ and $f(z)\!=\![\mathrm{Re}(z)\Vert\mathrm{Im}(z)]$. The map $f$ is an isometry from $\mathbb{C}^{N\times d}$ to $\mathbb{R}^{N\times 2d}$, so it does not change Dirichlet energy except for this real representation.

\noindent\textbf{Commuting case.} If $H$ commutes with $L$ (in particular $H=L$, or $H=A$ on regular graphs after normalization), $U_t$ and $L$ are simultaneously diagonalizable. Each non-trivial Laplacian coefficient is multiplied by a unit-modulus phase, hence
\begin{equation}\label{eq:specform}
E(f(U_t^nX^{(0)}))=\sum_{\ell=1}^{N}\mu_\ell\|\tilde X^{(0)}_\ell\|_2^2=E(X^{(0)}).
\end{equation}
By Lemma~\ref{lemma:spectralgapbound}, this constant is positive whenever the initial features have a non-constant component.

\noindent\textbf{Default $H=A$.} Now $H$ and $L$ need not commute, so pointwise invariance is not claimed. Instead, Lemma~\ref{lemma:spectralgapbound} gives $E(f(U_t^nX^{(0)}))\ge\Delta\|P_{\perp}U_t^nX^{(0)}\|_F^2$. Averaging over layers and applying (A3) yields
\begin{equation}
\liminf_{T\to\infty}\frac{1}{T}\sum_{n<T}E(f(U_t^nX^{(0)}))\ge\Delta\eta_0>0,
\end{equation}
where $\eta_0\!=\!\liminf_{T\to\infty}T^{-1}\sum_{n<T}\|P_{\perp}U_t^nX^{(0)}\|_F^2$. If $E(f(U_t^nX^{(0)}))\le ae^{-bn}$ for some $a,b>0$, its Ces\`aro average would converge to zero, contradicting the positive lower bound. Therefore the CTQW branch cannot exponentially over-smooth.
\end{proof}

\subsection{Spectral-Gap Lower Bound}
\label{proof:spectral_gap_bound}
\begin{lemma_app}\label{lemma:spectralgapbound}
For any $X\in\mathbb{C}^{N\times d}$, $E(X)\!\ge\!\Delta\|P_{\perp}X\|_F^2$. In the commuting CTQW case this gives $E(f(U_t^nX^{(0)}))\!\ge\!\Delta\|P_{\perp}X^{(0)}\|_F^2\!>\!0$ for every $n\!\ge\!0$.
\end{lemma_app}
\begin{proof}
Since $P_{\perp}$ removes only the zero eigenvector of $L$,
$E(X)=\sum_{\ell\ge 2}\mu_\ell\|\tilde X_\ell\|_2^2\ge\Delta\sum_{\ell\ge2}\|\tilde X_\ell\|_2^2=\Delta\|P_{\perp}X\|_F^2$.
If $H$ commutes with $L$, $\|P_{\perp}U_t^nX^{(0)}\|_F=\|P_{\perp}X^{(0)}\|_F$, and (A2) makes the bound strictly positive.
\end{proof}
Classical diffusion $e^{-Lt}$ admits the matching \emph{upper} bound $E(X^{(n+1)})\!\le\!e^{-2\Delta t}E(X^{(n)})$, decaying exponentially at rate $\Delta$. The same spectral quantity drives diffusion over-smoothing but yields a $\Delta$-linear floor for CTQW.

\subsection[Walk-Time and Effective Propagation Radius]{Walk-Time $t$ and Effective Propagation Radius}
\label{proof:propagation_radius}
\begin{proposition}\label{proposition:propagation}
Let $\|H\|_2\!=\!\lambda_{\max}$ and $d_{\mathcal{G}}(i,j)$ the geodesic distance. For any $t\!>\!0$,
\begin{equation}\label{eq:liebrobinson}
|(e^{-\mathrm{i}Ht})_{ij}|\;\le\;\exp(\lambda_{\max}t)\frac{(\lambda_{\max}t)^{d_{\mathcal{G}}(i,j)}}{d_{\mathcal{G}}(i,j)!}.
\end{equation}
Consequently, amplitudes beyond a radius linear in $\lambda_{\max}t$ are factorially small; the implementation uses the conservative screening radius $r_{\mathrm{eff}}(t)\!=\!\lceil 2\lambda_{\max}t/\pi\rceil$ and then applies the empirical $\epsilon$ threshold.
\end{proposition}
\begin{proof}
Expand $e^{-\mathrm{i}Ht}\!=\!\sum_{k\ge 0}(-\mathrm{i}t)^k H^k/k!$. Because $H$ is supported on graph edges, $(H^k)_{ij}=0$ whenever $k<d_{\mathcal{G}}(i,j)$. Thus only walks of length at least $d_{\mathcal{G}}(i,j)$ contribute. Bounding $|(H^k)_{ij}|\le\|H^k\|_2\le\lambda_{\max}^k$ gives
\begin{equation*}
|(e^{-\mathrm{i}Ht})_{ij}|\le\sum_{k\ge d_{\mathcal{G}}(i,j)}\frac{(\lambda_{\max}t)^k}{k!}\le\exp(\lambda_{\max}t)\frac{(\lambda_{\max}t)^{d_{\mathcal{G}}(i,j)}}{d_{\mathcal{G}}(i,j)!},
\end{equation*}
which is a graph Lieb--Robinson-type bound~\cite{liebrobinson1972}. Stirling's formula shows factorial decay once $d_{\mathcal{G}}(i,j)$ exceeds a constant multiple of $\lambda_{\max}t$, justifying a linear truncation radius. On a path, the exact amplitude is $|J_{|i-j|}(2t)|$~\cite{childs2002example}, whose ballistic front is also linear in $t$; this motivates the constant used by $r_{\mathrm{eff}}(t)$.
\end{proof}

\subsection{Krylov Implementation Error}
\label{proof:krylov}
\begin{proposition}\label{prop:krylov}
Let $\hat U_t^{(k)}$ denote the $k$-step Lanczos approximation of $e^{-\mathrm{i}Ht}$. For any $x\!\in\!\mathbb{C}^{N}$ and any $k\!\ge\!4\lambda_{\max}t$,
\begin{equation}\label{eq:krylov}
\|e^{-\mathrm{i}Ht}x-\hat U_t^{(k)}x\|_2\;\le\;12\,\|x\|_2\,\exp\!\Big(\!-\tfrac{k^2}{16\lambda_{\max}t}\Big).
\end{equation}
Hence the discretised CTQW branch satisfies
\begin{equation}\label{eq:krylovfloor}
E_{CTQW}^{\mathrm{approx}}(X^{(n)})\;\ge\;E_{CTQW}^{\mathrm{exact}}(X^{(n)})-\epsilon_{\mathrm{kry}},
\end{equation}
with $\epsilon_{\mathrm{kry}}\!\le\!144\,\mu_{\max}B^2\,\exp\!\big(\!-k^2/(8\lambda_{\max}t)\big)$ whenever the layer input satisfies $\|X^{(n)}\|_F\le B$.
\end{proposition}
\begin{proof}[Sketch]
Eq.~\ref{eq:krylov} is Theorem~4 of~\citet{hochbruck1997krylov} specialised to the skew-Hermitian generator $-\mathrm{i}H$ of spectral radius $\lambda_{\max}$; constants $12,16$ track through their proof. For Eq.~\ref{eq:krylovfloor}, write the one-application error as $\Xi$ with $\|\Xi\|_F\!\le\!12B e^{-k^2/(16\lambda_{\max}t)}$. Since $L\preceq\mu_{\max}I$, $|E(X+\Xi)-E(X)|\le\mu_{\max}\|\Xi\|_F^2+2\mu_{\max}\|\Xi\|_F\|X\|_F$, yielding the stated perturbation bound after absorbing constants. Thus Krylov approximation can only reduce the CTQW energy floor by a controllable, exponentially small term.
\end{proof}
At $k\!=\!20$ and the selected $t$ range, the exponential term is negligible in all benchmarks; the implementation therefore preserves the qualitative non-decay guarantee up to numerical precision.

\subsection{All-Pass Spectral Response of CTQW-based Aggregation}
\label{proof:allpass}
\begin{proposition}\label{prop:allpass}
Let $H$ admit the spectral decomposition $H\!=\!U\Lambda U^{\dagger}$ with real eigenvalues $\{\lambda_l\}_{l=1}^{N}$. For any input $X\!\in\!\mathbb{C}^{N\times d}$, write $X\!=\!\sum_l u_l\alpha_l^{\top}$ with $\alpha_l\!\in\!\mathbb{C}^d$ the spectral coefficients. Then the CTQW propagator preserves the magnitude of every spectral component:
\begin{equation}\label{eq:allpass}
\|(\,e^{-\mathrm{i}Ht}X\,)_{\text{mode }l}\|_2\;=\;|e^{-\mathrm{i}\lambda_l t}|\,\|\alpha_l\|_2\;=\;\|\alpha_l\|_2\qquad\forall\,l\!=\!1,\ldots,N.
\end{equation}
Hence CTQW-based Aggregation realises an exact all-pass filter on the eigenbasis of $H$.
\end{proposition}
\begin{proof}
Expand $X\!=\!\sum_l u_l\alpha_l^{\top}$. Since $\{u_l\}$ is orthonormal and $e^{-\mathrm{i}Ht}u_l\!=\!e^{-\mathrm{i}\lambda_l t}u_l$,
$e^{-\mathrm{i}Ht}X\!=\!\sum_l e^{-\mathrm{i}\lambda_l t}\,u_l\alpha_l^{\top}$, so the $l$-th mode coefficient becomes $e^{-\mathrm{i}\lambda_l t}\alpha_l$. As $\lambda_l\!\in\!\mathbb{R}$, $|e^{-\mathrm{i}\lambda_l t}|\!=\!1$, giving Eq.~\ref{eq:allpass}. In contrast, classical diffusion $e^{-Lt}$ multiplies the $l$-th coefficient by $e^{-\lambda_l t}\!<\!1$ for $\lambda_l\!>\!0$, attenuating every non-trivial mode at rate $\ge\!e^{-\Delta t}$ per layer; this strictly contradicts the all-pass property of CTQW.
\end{proof}
Proposition~\ref{prop:allpass} formalises the magnitude-preserving filter claim made in the method section: no frequency component is damped, which is precisely the structural reason CTQW-based Aggregation supplies the energy floor used in Proposition~\ref{proposition:oversmoothing_CTQW}.

\subsection{Branch-Mixing Floor}
\label{proof:mixing}
\begin{lemma_app}\label{lemma:mixing}
Let $\mathcal{S}^{(n)}$ be the set of normalized full edge differences $\delta=[\delta_{LF},\delta_{CTQW},\delta_{Att}]$ at layer $n$. Assume the learned mixer has conditional CTQW gain $c_{CTQW}>0$ on this subspace, meaning $\|\sigma(\delta W_\theta)\|_2\ge c_{CTQW}\|\delta_{CTQW}\|_2$ for all $\delta\in\mathcal{S}^{(n)}$. Then
\begin{equation}
E(X^{(n+1)})\;\ge\;c_{CTQW}^2\,E_{CTQW}^{(n)}.
\end{equation}
\end{lemma_app}
\begin{proof}
Apply the gain assumption to each normalized edge difference of the CTQW block and sum over edges. This rules out the only degenerate failure mode in which the learned projection intentionally maps all CTQW differences to zero. Combining this lemma with Proposition~\ref{proposition:oversmoothing_CTQW} shows that a retained CTQW branch prevents exponential decay of the mixed representation.
\end{proof}

\bibliographystyle{ACM-Reference-Format}
\bibliography{sample-base}










\end{document}